\documentclass[journal]{IEEEtran}

\ifCLASSINFOpdf
\else
\fi

\usepackage{cite}
\usepackage[hidelinks]{hyperref}
\usepackage[capitalize]{cleveref}
\usepackage{times}
\usepackage{epsfig}
\usepackage{graphicx}
\usepackage[cmex10]{amsmath}
\usepackage{amssymb}
\usepackage{color}
\usepackage{xcolor}
\usepackage{multirow}
\usepackage{array}
\usepackage{epstopdf}
\usepackage{subfig}
\usepackage{caption}
\usepackage[font=footnotesize]{caption}
\usepackage{booktabs}
\usepackage{enumerate}
\usepackage{float}
\usepackage{adjustbox}
\usepackage{bm}
\usepackage{upgreek}
\usepackage[super]{nth}
\usepackage{marvosym}
\usepackage{textcomp }
\usepackage {tabularx}
\usepackage{eso-pic}

\newcommand{\etal}{\textit{et al}. }

\begin{document}
\AddToShipoutPictureBG*{%
    \AtPageUpperLeft{%
        \setlength\unitlength{1in}%
        \hspace*{\dimexpr0.5\paperwidth\relax}
        \makebox(0,-0.25)[c]{\small This work has been submitted to the IEEE for possible publication.}
        \makebox(0,-0.5)[c]{\small Copyright may be transferred without notice, after which this version may no longer be accessible.}%
    }}

\title{Channel-wise and Spatial Feature Modulation Network for Single Image Super-Resolution}

\author{Yanting~Hu,
              Jie~Li,
             Yuanfei~Huang,
             and ~Xinbo~Gao,~\IEEEmembership{Senior Member,~IEEE,} \\
             \{yantinghu2012\}@gmail.com, \{leejie, xbgao\}@mail.xidian.edu.cn, \{yf\_huang\}@stu.xidian.edu.cn
}

\maketitle

\begin{abstract}
The performance of single image super-resolution has achieved significant improvement by utilizing deep convolutional neural networks (CNNs). The features in deep CNN contain different types of information which make different contributions to image reconstruction. However, most CNN-based models lack discriminative ability for different types of information and deal with them equally, which results in the representational capacity of the models being limited. On the other hand, as the depth of neural networks grows, the long-term information coming from preceding layers is easy to be weaken or lost in late layers, which is adverse to super-resolving image. To capture more informative features and maintain long-term information for image super-resolution, we propose a channel-wise and spatial feature modulation (CSFM) network in which a sequence of feature-modulation memory (FMM) modules is cascaded with a densely connected structure to transform low-resolution features to high informative features. In each FMM module, we construct a set of channel-wise and spatial attention residual (CSAR) blocks and stack them in a chain structure to dynamically modulate multi-level features in a global-and-local manner. This feature modulation strategy enables the high contribution information to be enhanced and the redundant information to be suppressed. Meanwhile, for long-term information persistence, a gated fusion (GF) node is attached at the end of  the FMM module to adaptively fuse hierarchical features and distill more effective information via the dense skip connections and the gating mechanism. Extensive quantitative and qualitative evaluations on benchmark datasets illustrate the superiority of our proposed method over the state-of-the-art methods.
\end{abstract}


\IEEEpeerreviewmaketitle

\section{Introduction}

\IEEEPARstart{S}{ingle} image super-resolution (SISR), which aims at reconstructing a high-resolution (HR) image from its single low-resolution (LR) counterpart, is an ill-posed inverse problem. To tackle such an inverse problem, numerous learning-based super-resolution (SR) methods have been proposed to learn the mapping function between LR and HR image pairs via probabilistic graphical model \cite{1Freeman2000IJCV, 2Polatkan2015TPAMI}, neighbor embedding \cite{3Chang2004CVPR, 4Jiang2016TCSVT}, sparse coding \cite{5Yang2010TIP_Sparse, 6He2013CVPR}, linear or nonlinear regression \cite{7Timofte2014ACCV_A+, 8Hu2016TIP_SERF, Huang2017TCSVT},  and random forest \cite{9Schulter2015CVPR_SRforests}.

\begin{figure}[t]
\vspace{-0.15cm}
		\captionsetup[subfloat]{labelformat=empty}
		\begin{center}
			\setlength\tabcolsep{0.02cm}
			\begin{tabular}[b]{c c c}
                    \multicolumn{2}{c}{\multirow{2}{*}[1.98cm]{
						\subfloat[``{\em img093}'' from Urban100~\cite{44Huang2015CVPRSelfExp}]
						{\includegraphics[height=0.53\linewidth,  trim={0.7cm  0cm  1.5cm 0cm}, clip]
							{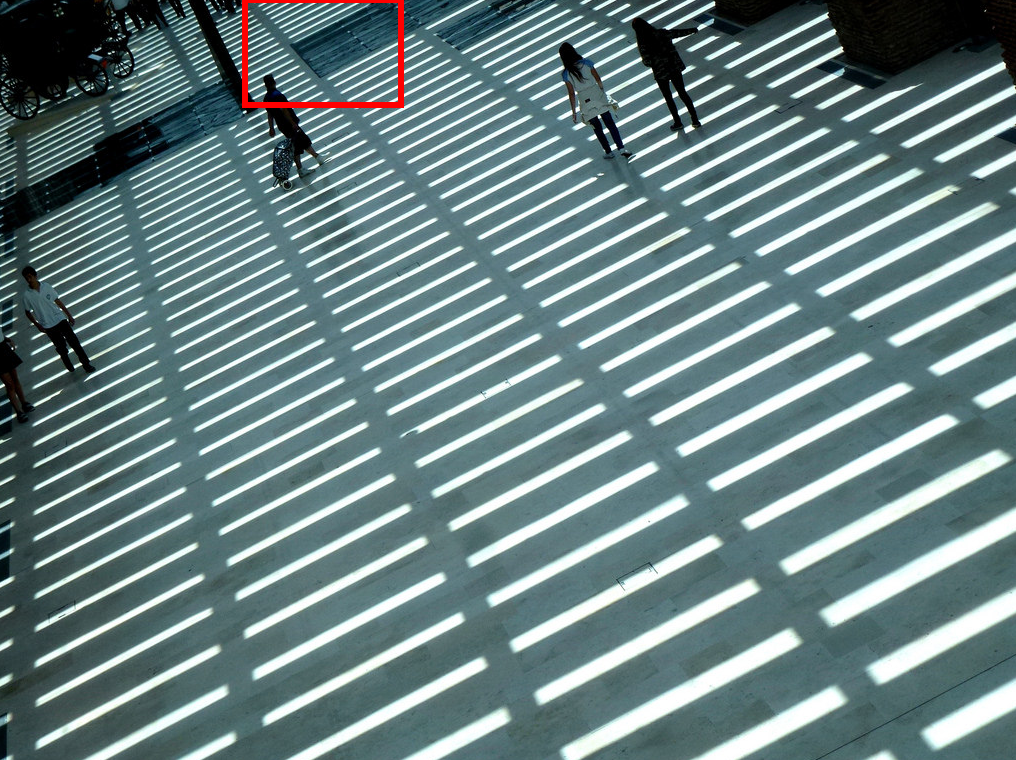}}}} &
				\subfloat[\centerline{Ground Truth} \protect  \linebreak \centerline{PSNR / SSIM} ]{
					\includegraphics[height=0.22\linewidth]
					{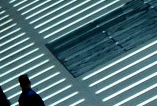}} \\ [-0.3cm]& &
				\subfloat[Bicubic\protect \linebreak 23.63dB / 0.8041]{
					\includegraphics[height=0.22\linewidth]
					{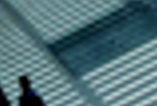}} \\ [-0.3cm]
				
				\subfloat[\centerline{EDSR~\cite{20Lim2017CVPRW_EDSR}} \protect\linebreak \centerline{29.56dB / 0.9336}]{
					\includegraphics[ height=0.22\linewidth]
					{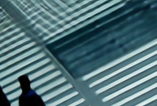}} &
				\subfloat[\centerline{RDN~\cite{24Zhang2018CVPR_RDN}} \protect\linebreak \centerline{28.59dB / 0.9286}]{
					\includegraphics[height=0.22\linewidth]
					{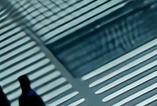}}	&
				\subfloat[\centerline{\textbf{CSFM}(ours)} \protect \linebreak \centerline{\bf {32.06dB} / \bf {0.9462}}]{ 
					\includegraphics[ height=0.22\linewidth]
					{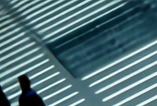}}
			\end{tabular}		
		\end{center}
		\setlength{\abovecaptionskip}{0cm}
		\captionsetup{justification=raggedright,singlelinecheck=false, width=0.99\linewidth,  belowskip=-8pt}
		\caption{The comparisons of our proposed method (CSFM)  with existing methods on single image super-resolution for a scale factor of $4\times$. Our proposed CSFM network generates more realistic visual result. }
   \label{fig:Subjective comparisons: CSFM EDSR RDN}		
\end{figure}

\begin{figure*}
\vspace{-0.2cm}
\captionsetup[subfigure]{farskip = 0pt}
\captionsetup{belowskip=-14pt}
    \small
    \begin{minipage}[b]{1\linewidth}
        \centering
        \subfloat[The architecture of the proposed CSFM network]{
            \centering
            \includegraphics[width=1\linewidth]{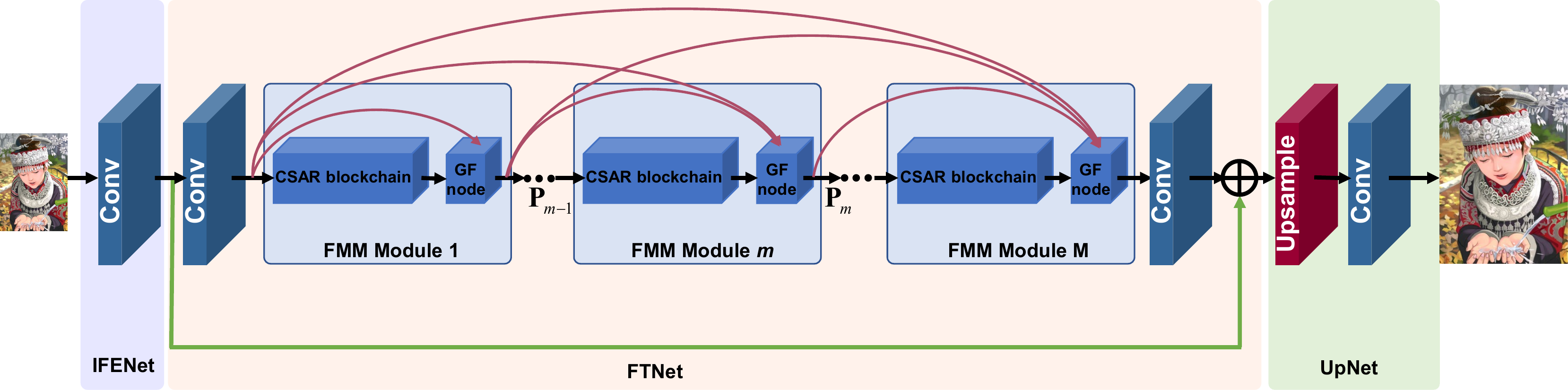}
        }\\
        \vspace{0.2cm}
        \subfloat[The structure of feature-modulation memory (FMM) module in CSFM network]{
            \centering
            \includegraphics[width=1\linewidth]{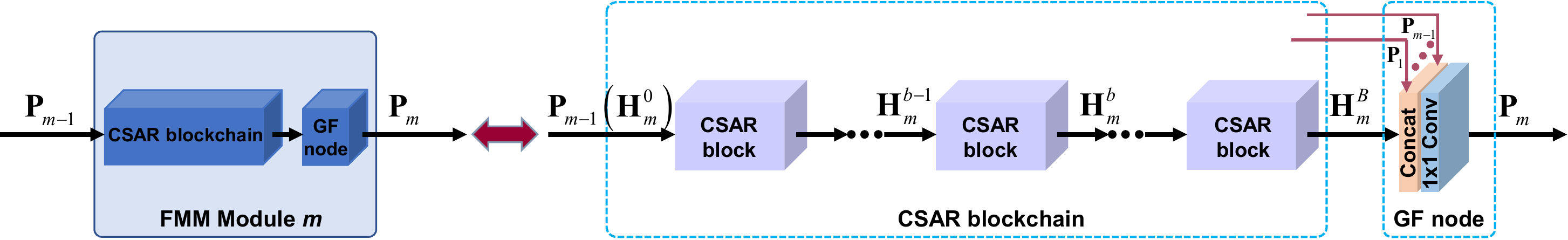}
        }
        \end{minipage}
\caption{ The architecture of our CSFM network and the structure of FMM module in CSFM network. (a) The overall architecture of the proposed CSFM network, which adopts adaptive feature-modulation strategy, long-term information persistence mechanism and post-upscaling scheme to boost SR performance. (b) The feature-modulation memory (FMM) module in (a), which exploits a chain of channel-wise and spatial attention residual (CSAR) blocks to capture more informative features and utilizes the gated fusion (GF) node to fusion long-term information from  the preceding FMM modules and short-term information from the current module.}
\label{fig:The architectures}
\end{figure*}

More recently, benefiting from the powerful representational ability of convolutional neural networks (CNNs), deep-learning-based SR methods have achieved better performances in terms of effectiveness and efficiency. As an early first attempt, SRCNN \cite{10Dong2016TPAMI}  proposed by Dong \etal employed three convolutional layers to predict the nonlinear mapping function from bicubic upscaled middle resolution image to high resolution image, which outperformed most conventional SR methods. Later, various works followed the similar network design and consistently improved SR performance via residual learning \cite{11KimJ2016CVPR_VDSR,  12TaiY2017CVPR_DRRN}, recursive learning \cite{12TaiY2017CVPR_DRRN, 13KimJ2016CVPR_DRCN}, symmetric skip connections \cite{14Mao2016NIPS_RED} and cascading memory blocks \cite{15TaiY2017ICCV_MemNet}. Differing from the above pre-upscaling approaches which operated SR on bicubic upsampled images, FSRCNN \cite{16Dong2016ECCV_FSRCNN} and ESPCN \cite{17Shi2016CVPR_Sub-Pixel}, designed by Dong \etal  and Shi \etal respectively, extracted features from the original LR images and upsampled spatial resolution only at the end of the processing pipeline via a deconvolution layer or a sub-pixel convolution module \cite{17Shi2016CVPR_Sub-Pixel}. Following this post-upscaling architecture, Ledig \etal  \cite{18LedigC2017CVPR_SRResNet} employed the residual blocks proposed in \cite{19He2016CVPR_ResnetIR} to construct a deeper network (SRResnet) for image SR, which was further improved by EDSR \cite{20Lim2017CVPRW_EDSR} and MDSR \cite{20Lim2017CVPRW_EDSR} via removing unnecessary modules. Further, to conveniently pass information across several layers, dense blocks \cite{22Huang2017CVPR_DenseNet} were also introduced to construct several deep networks \cite{23Tong2017ICCV_DenseSR, 24Zhang2018CVPR_RDN, 25Wang2018CVPRW_ProSR} for suiting image super-resolution. Meanwhile, to simplify the difficulty of direct super-resolving the details, \cite{25Wang2018CVPRW_ProSR, 26LaiWS2017CVPR_LapSRN, He2018TCSVT} adopted the progressive structure to reconstruct HR image in a stage-by-stage upscaling manner. In addition, \cite{27Haris2018CVPR_DBPN, 28Han2018arXiv_DSRN} incorporated the feedback mechanism into network designs for exploiting both LR and HR signals jointly.

Although these existing deep-learning-based approaches have made good efforts to improve SR performance, the reconstruction of high frequency details for SISR is still a challenge. In deep neural networks, the LR inputs and extracted features contain different types of information across channels, spaces and layers, such as low-frequency and high-frequency information or low-level and high-level features, which have different reconstruction difficulties (e.g., the high-frequency features or the pixels on the texture areas are more difficult to reconstruction than the low-frequency features or the pixels on the flat areas) as well as different contributions to recovering the implicit high-frequency details. However, the most CNN-based methods consider different types of information equally and lack flexible modulation ability in dealing with them, which resultantly limits the representational ability and fitting capacity of the deep networks. Therefore,  for the deeper neural networks,  simply increasing depth or width  can hardly achieve better improvement. On the other hand, for image restoration tasks, the hierarchical features produced by deep neural networks are informative and useful. However, many very deep networks, such as VDSR \cite{11KimJ2016CVPR_VDSR}, LapSRN \cite{26LaiWS2017CVPR_LapSRN}, EDSR \cite{20Lim2017CVPRW_EDSR} and IDN \cite{30Hui2018CVPR_IDN}, adopt single-path direct connections or short skip connections among layers, where hierarchical features could hardly be fully utilized and long-term information that provides some clues for SR would be lost as the network depth grows. Although SRDenseNet \cite{23Tong2017ICCV_DenseSR} and RDN \cite{24Zhang2018CVPR_RDN} employ dense-connection blocks for SR to fuse different levels of features, the extreme connectivity pattern in their networks not only hinders their scalability to large width or high depth but also produces redundant computation. Memory blocks adopted in MemNet \cite{15TaiY2017ICCV_MemNet} also integrate information from the preceding memory blocks to achieve persistent memory, but the fused features are extracted from bicubic pre-upscaled images which might lose some details and produce new noises. Therefore, how to effectively make full use multi-level, channel-wise and spatial features within neural  networks is crucial for HR image reconstruction and remains to be explored.

To address these issues, we propose a Channel-wise and Spatial Feature Modulation network (illustrated in Fig.~\ref{fig:The architectures}) for SISR, named CSFM, which not only adaptively learns to pay attention to every feature entry in the multi-level, channel-wise and spatial feature responses but also fully and effectively exploits the hierarchical features to maintain persistent memory. In the CSFM network, we construct a feature-modulation memory (FMM) module (shown in Fig.~\ref{fig:The architectures}(b)) as the building module and stack several FMM modules with a densely connected structure. An FMM module contains a channel-wise and spatial attention residual (CSAR) blockchain and a gated fusion (GF) node. In the CSAR blockchain, we develop a channel-wise and spatial attention residual (CSAR) block via integrating the channel-wise and spatial attentions into the residual block \cite{19He2016CVPR_ResnetIR} and stack a collection of CSAR blocks to modulate multi-level features  for adaptively capturing more important information. In addition, by adopting a GF node in the FMM module, the states of the current FMM module and of the preceding FMM modules are conveniently concatenated and adaptively fused for short-term and long-term information preservation as well as for information flow enhancement. As shown in Fig.~\ref{fig:Subjective comparisons: CSFM EDSR RDN}, our proposed CSFM network generates more realistic visual result compared with other methods.

In summary, the major contributions of our proposed SISR method are three-fold:

1). We develop a CSAR block via combining channel-wise and spatial attention mechanisms into the residual block, which can adaptively recalibrate the feature responses in a global-and-local manner by explicitly modelling channel-wise and spatial feature interdependencies.

2). We construct an FMM module via stacking a set of CSAR blocks to modulate multi-level features and adding a GF node to adaptively fuse hierarchical features for important information preservation. The  block-stacking structure in the FMM module enables it to capture different types of attention and then enhance high contribution information for image super-resolution, while the gating mechanism help it to adaptively distill more effective information from short-term and long-term states.

3) We design a CSFM network  for accurate single image SR, in which the stacked FMM modules enhance discriminative learning ability of the network and the densely connected structure helps to fully exploit multi-level information as well as ensures maximum information flow between modules.

The remainder of this paper is organized as follows. Section II discusses the related  SISR methods  and  correlative mechanisms applied in neural networks. Section III describes the proposed CSFM network for SR in detail. Model analysis and experimental comparisons with other state-of-the-art methods are presented in Section IV, and Section V concludes the paper with observations and discussions.

\section{Related Work}
Numerous SISR methods, different learning mechanisms and various network architectures have been proposed in the literatures. Here, we focus our discussions on the approaches which are related to our method.

\subsection{Deep-learning based Image Super-Resolution}
Since Dong \etal  [10] first proposed a super-resolution convolutional neural network (SRCNN) to predict the nonlinear relationship between bicubic upscaled image and HR image, various CNN architectures have been studied for SR. As deeper CNNs have larger receptive fields to capture more contextual information, Kim \etal proposed two deep networks of VDSR \cite{11KimJ2016CVPR_VDSR} and DRCN \cite{13KimJ2016CVPR_DRCN}  which utilized global residual learning and recursive layers respectively to improve SR accuracy. To control the number of model parameters and maintain persistent memory, Tai \etal constructed the recursive blocks with global-and-local residual learning in DRRN \cite{12TaiY2017CVPR_DRRN} and designed the memory blocks with dense connections in MemNet \cite{15TaiY2017ICCV_MemNet}. For these methods, the LR images need be bicubic interpolated to the desired size before entering the networks, which inevitably increases the computational complexity and might produce new noise.

For alleviating the computational loads and overcoming the disadvantage of the pre-upscaling structure, Dong \etal \cite{16Dong2016ECCV_FSRCNN} exploited the deconvolution operator to upscale spatial resolution at the network tail. Later, Shi \etal \cite{17Shi2016CVPR_Sub-Pixel}  proposed a more effective sub-pixel convolution layer to replace the deconvolution layer for upscaling the final LR feature-maps into the HR output, which was recently extended by an enhanced upscaling module (EUM) \cite{29Kim2018CVPRW_EUSR} via applying residual learning and multi-path concatenation into the  module. Benefiting from this post-upscaling strategy, more and more deeper networks, such as SRResnet \cite{18LedigC2017CVPR_SRResNet}, EDSR \cite{20Lim2017CVPRW_EDSR}  and SRDenseNet \cite{23Tong2017ICCV_DenseSR}, achieved high performances with less computational load. Recently, Hui \etal \cite{30Hui2018CVPR_IDN}  developed the information distillation blocks and stacked them to construct a deep and compact convolutional network. And, Zhang \etal \cite{24Zhang2018CVPR_RDN} proposed a residual dense network (RDN) which used the densely connected convolutional layers to extract abundant local features and adopted the local-and-global feature fusion procedure to adaptively fuse hierarchical features in the LR space.

Taking the effectiveness of post-upscaling strategy into account, we also apply the sub-pixel convolution layer \cite{17Shi2016CVPR_Sub-Pixel} at the end of network for upscaling spatial resolution. Furthermore, we exploit the feature modulation mechanism to enhance the discriminative ability of  the network for different types of information.

\subsection{Attention Mechanism}
The aim of attention mechanism in neural network is to recalibrate the feature responses towards the most informative and important components of the inputs. Recently, some works have focused on the integration of attention modules within deep network architectures on a range of tasks, such as image generation \cite{31Mansimov2016ICLR_CaptionAtten}, image captioning \cite{32Xu2015ICML_CaptionGenera, 33Chen2017CVPR_SCA-CNN}, image classification \cite{34Hu2018CVPR_SEnet, 35Wang2017CVPR_ResidualAttenClass} and image restoration \cite{21Zhang2018ECCV_RCAN, 36Wang2018CVPR_SFTSR}. Xu \etal \cite{32Xu2015ICML_CaptionGenera} proposed a visual attention model for image captioning, which used ¡°hard¡± pooling to select the most probably attentive region or ¡°soft¡± pooling to average the spatial features with attentive weights. Xu \etal \cite{37Xu2016ECCV_QuestAnswer} further refined the spatial attention model by stacking two spatial attention models for visual question answering. Moreover, by investigating the interdependencies between the channels of the convolutional features in a network, Hu \etal \cite{34Hu2018CVPR_SEnet} introduced a channel-wise attention mechanism and proposed a squeeze-and-excitation (SE) block to adaptively recalibrate channel-wise feature responses for image classification. Recently, inspired by SE networks, Zhang \etal \cite{21Zhang2018ECCV_RCAN} integrated the channel-wise attention into the residual blocks and proposed a very deep residual channel attention network which pushed the state-of-the-art performance of SISR forward. In addition, Chen \etal \cite{33Chen2017CVPR_SCA-CNN} stacked the spatial and channel-wise attention modules at multiple layers for image captioning, where the second attention (spatial attention or channel-wise attention) was operated on the attentive feature-maps recalibrated by the first one (channel-wise attention or spatial attention). Besides the spatial and channel-wise attentions, Wang \etal \cite{36Wang2018CVPR_SFTSR} utilized semantic segmentation probability maps as prior knowledge and introduced semantic attention to modulate spatial features for realistic texture generation. However, this model requires external resources to train these semantic attributes.

Inspired by attention mechanism and considering that there are different types of information within and across feature-maps which have different contributions for image SR, we  combine channel-wise and spatial attentions into the residual blocks to adaptively modulate feature representations in  a global-and-local way for capturing more important information.

\begin{figure}[t]
\captionsetup[subfigure]{farskip = 0pt}
\captionsetup{belowskip=-20pt}
    \small
    \begin{minipage}[b]{1\linewidth}
        \centering
        \subfloat[Channel-wise and spatial attention residual (CSAR) block]{
            \centering
            \includegraphics[width=1\linewidth]{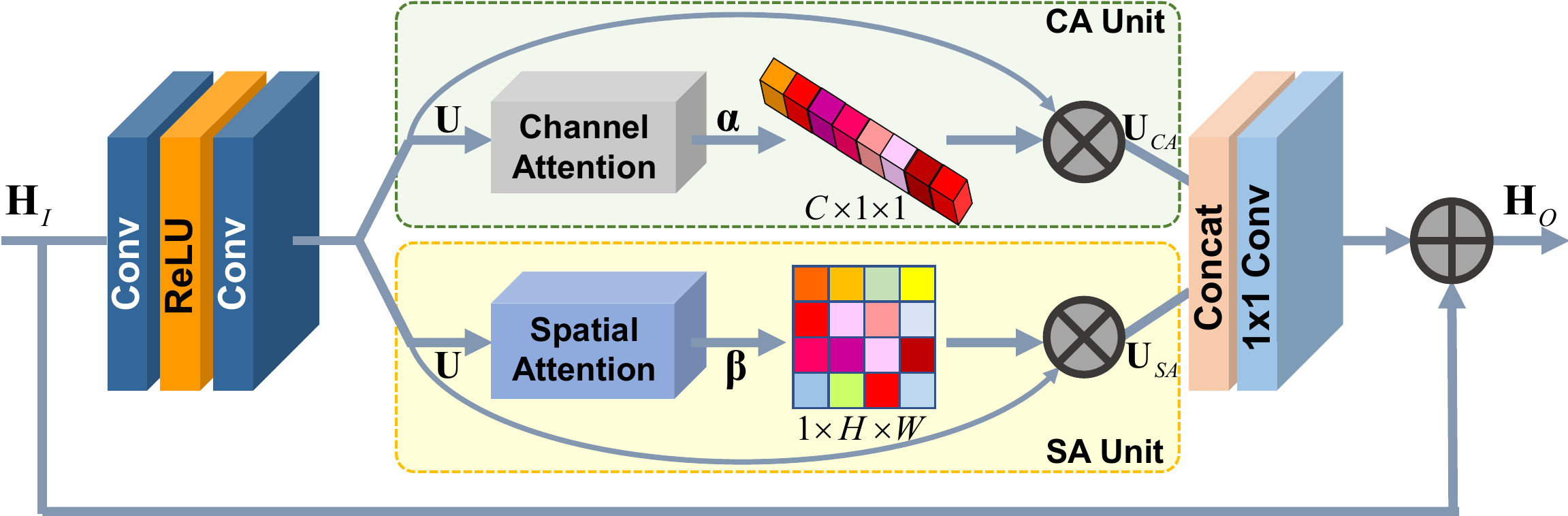}
        }\\
        \subfloat[The operations of channel-wise attention]{
            \centering
            \includegraphics[width=0.8\linewidth]{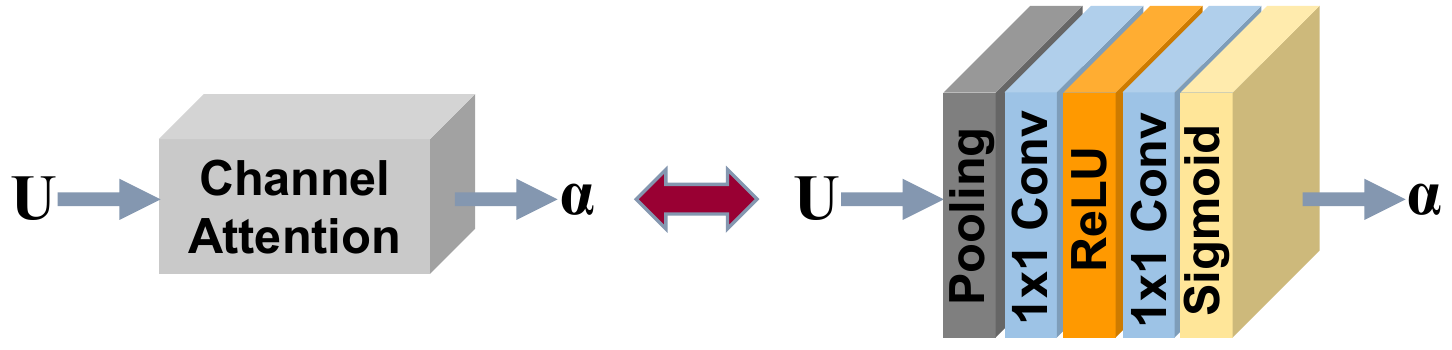}
        }\\
         \subfloat[The operations of spatial attention]{
            \centering
            \includegraphics[width=0.8\linewidth]{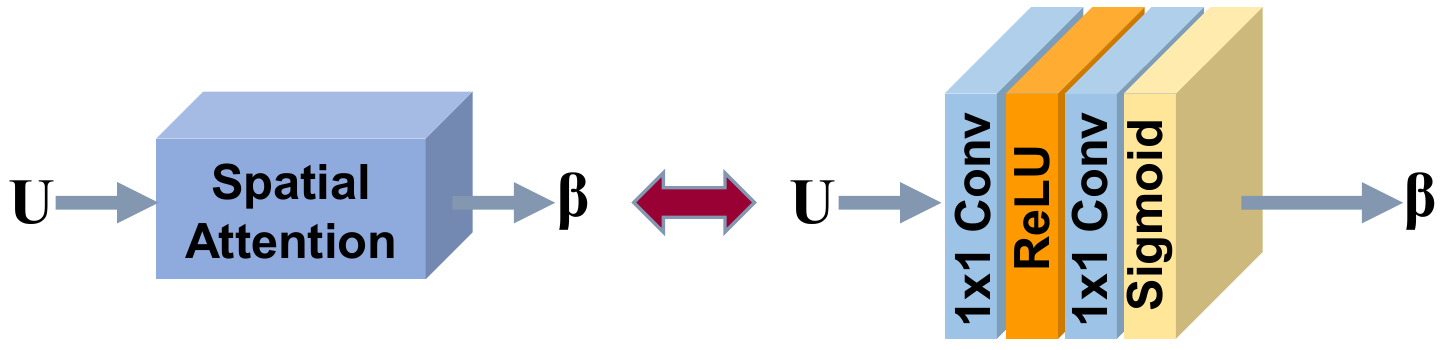}
        }
        \end{minipage}
\caption{The diagram of channel-wise and spatial attention residual (CSAR) block, where $ \bigotimes$ denotes element-wise product. (a) The CSAR block, which integrates the channel-wise attention and spatial attention into the residual block to modulate the residual features. (b) The operations of channel-wise attention, including global-pooling, convolutions and activations, by which the channel-wise attention weights are calculated. (c) The operations of convolutions and activations in spatial attention by which the spatial attention mask is generated.}
\label{fig:The CSAR block}
\vspace{0.3cm}
\end{figure}

\subsection{Skip Connections}
As the depth of a network grows, the problems of information flow weakened and gradient vanishing hamper the training of  the network. Many recent methods have been devoted to resolving these problems. ResNets proposed by He \etal \cite{19He2016CVPR_ResnetIR} was built by stacking a sequence of residual blocks,  which utilized the skip connections between layers to improve information flow and make training easier. The residual blocks were also widely applied in \cite{18LedigC2017CVPR_SRResNet, 20Lim2017CVPRW_EDSR} to construct very wide and deep networks for SR performance improvement. To fully explore the advantages of skip connections, Huang \etal \cite{22Huang2017CVPR_DenseNet}  constructed DenseNets by directly connecting each layer to all previous layers. Meanwhile, in order to make the networks scale to deep and wide ones, block compression was applied in DenseNets to halve the number of channels in the concatenation of previous layers. The dense connections were  utilized in \cite{15TaiY2017ICCV_MemNet,  23Tong2017ICCV_DenseSR, 24Zhang2018CVPR_RDN} for image SR to improve the flows of information and gradient throughout the networks as well. However, the extremely dense connections and frequent concatenations may increase information redundancy and computational cost. Considering these, Chen \etal \cite{38Chen2017NIPS_DualPN}  combined the insights of ResNets \cite{19He2016CVPR_ResnetIR} and DenseNets \cite{22Huang2017CVPR_DenseNet} and proposed a DualPathNet which utilized both concatenation and summation for previous features.

Recognizing both advantages of residual path in residual block and densely connected paths in dense block, we stack several attention-based residual blocks within each module and utilize the densely connected paths between modules for effective feature re-exploitation and important information preservation.

\section{The Proposed CSFM Network}
The proposed CSFM network for SISR, outlined in Fig.~\ref{fig:The architectures}, consists of an initial feature extraction sub-network (IFENet), a feature transformation sub-network (FTNet) and an upscaling sub-network (UpNet). The IFENet is applied to represent a LR input as a set of feature-maps via a convolutional layer. The FTNet is designed to capture more informative features for SR by a sequence of stacked feature-modulation memory (FMM) modules and two convolutional layers. The transformed features are then fed into the UpNet to generate the HR image. In this section, we detail the proposed model, from the  channel-wise and spatial attention residual (CSAR) block to the FMM module and finally the overall network architecture.

\subsection{The CSAR Block}
The features generated by a deep network contain different types of information across channels and spatial regions which have different contributions for the high-frequency details recovery. If we are able to increase the network's sensitivity to higher contribution features and make it focus on learning more important features, the representational power of the network would be enhanced and the performance improved. Keeping that in mind, we design a channel-wise attention (CA) unit  and a spatial attention (SA) unit by utilizing the interdependencies between channels and spatial locations of the features,  and then combine two types of attention into the residual blocks to adaptively modulate feature representations.

\subsubsection{The CA Unit}
The aim of the CA unit is to perform feature recalibration in a global way where the per-channel ¡°summary statistics¡± are calculated and then used to selectively emphasis informative feature-maps as well as suppress useless ones (e.g. redundant feature-maps). The structure of the CA unit is illustrated in Fig.~\ref{fig:The CSAR block}(a)--(b). We denote $\mathbf{U}=\left[ {{\mathbf{u}}_{1}},{{\mathbf{u}}_{2}},\cdots ,{{\mathbf{u}}_{C}} \right]$ as the input of the CA unit, which consists of  $C$  feature-maps with size of  $H\times W$. To generate channel-wise ¡°summary statistics¡±  $\mathbf{z}\in {{\mathbb{R}}^{C\times 1\times 1}}$,   the global average pooling is operated on individual feature channels across spatial dimensions $H\times W$,  as done in  \cite{34Hu2018CVPR_SEnet}. The $c\text{-th}$  element of   $\mathbf{z} $  is computed by
\begin{equation}
\label{Eq: eq 1}
{{z}_{c}}=\frac{1}{H\times W}\sum\limits_{i=1}^{H}{\sum\limits_{j=1}^{W}{{{\mathbf{u}}_{c}}(i,j)}},
\end{equation}
where ${{\mathbf{u}}_{c}}(i,j)$ is the value at position $\left( i,j \right)$ of the $c\text{-th}$ channel ${{\mathbf{u}}_{c}}$. To assign different attentions to different types of feature-maps, we employ a gating mechanism with a sigmoid activation to summary statistic  $\mathbf{z}$.  The process is represented as follows.
\begin{equation}
\label{Eq: eq 2}
\bm{\upalpha }=\sigma (\mathbf{W}_{CA}^{2}*\delta (\mathbf{W}_{CA}^{1}*\mathbf{z}+\mathbf{b}_{CA}^{1})+\mathbf{b}_{CA}^{2}),
\end{equation}
where $\sigma \left( \cdot  \right)$ and  $\delta \left( \cdot  \right)$ represent the sigmoid and ReLU \cite{39Nair2010ICML_ReLU} functions respectively, and $*$ denotes the convolution operation. $\mathbf{W}_{CA}^{1}\in {{\mathbb{R}}^{\frac{C}{r}\times C\times 1\times 1}}$ and $\mathbf{b}_{CA}^{1}\in {{\mathbb{R}}^{\frac{C}{r}}}$ are the weights and bias in the first convolutional layer which is followed by ReLU activation and used to decrease the number of channels of  $\mathbf{z} $  by the reduction ratio $r$. Next, the number of channels is increased back to the original amount via another convolutional layer with parameters of $\mathbf{W}_{CA}^{2}\in {{\mathbb{R}}^{C\times \frac{C}{r}\times 1\times 1}}$ and $\mathbf{b}_{CA}^{2}\in {{\mathbb{R}}^{C}}$. In addition, the channel-wise attention weights $\bm{\upalpha }\in {{\mathbb{R}}^{C\times 1\times 1}}$ are adapted to the values between $0$ and $1$ by sigmoid function $\sigma \left( \cdot  \right)$, and then used to rescale the input features as follows.
\begin{equation}
\label{Eq: eq 3}
{{\mathbf{U}}_{CA}}={{\Phi }_{CA}}(\mathbf{U})={{f}_{CA}}(\mathbf{U}, \bm{\upalpha }),
\end{equation}
where ${{f}_{CA}}\left( \cdot  \right)$  is a channel-wise multiplication for feature channels and corresponding channel weights, ${\mathbf{U}}_{CA}$ is the channel-wise recalibrated output,  and ${{\Phi }_{CA}}(\cdot )$ represents the CA unit which is apparently conditioned on the input $\mathbf{U}$.

With the above process, the CA unit is able to adaptively modulate the channel-wise features according to the channel-wise statistics of input, and help the network boost the channel-wise feature discriminability.

\subsubsection{The SA Unit}
The channel-wise attention exploits global average pooling to squeeze global spatial information into a channel statistical descriptor, by which the spatial information within each feature-map is yet removed. On the other hand, the information contained in the inputs and feature-maps is also diverse over spatial positions. For example, the edge or texture regions usually contain more high-frequency information while the smooth areas have more low-frequency information. Therefore, to recover high-frequency details for image SR, it is helpful to make the network have discriminative ability for different local regions and pay more attentions to the regions which are more important and more difficult to reconstruct.

Considering aforementioned discussion, besides the channel-wise attention, we explore a complementary form of attention termed as spatial attention to improve the representations of the network. As shown in  Fig.~\ref{fig:The CSAR block}(a)--(c),  let $\mathbf{U}=\left[ {{\mathbf{u}}_{1}},{{\mathbf{u}}_{2}},\cdots ,{{\mathbf{u}}_{C}} \right]$ be an input for the SA unit, which has $C$  feature-maps with size of $H\times W$.  To make use of feature channel interdependencies of the input and inspired by the local computations in computational-neuroscience models \cite{40Itti2001Neurosci}, we use a two-layers neural network followed by a sigmoid function to generate a spatial attention mask $\bm{\upbeta }\in {{\mathbb{R}}^{1\times H\times W}}$. Below is the definition of the SA unit.
\begin{equation}
\label{Eq: eq 4}
\bm{\upbeta }=\sigma (\mathbf{W}_{SA}^{2}*\delta (\mathbf{W}_{SA}^{1}*\mathbf{U}+\mathbf{b}_{SA}^{1})+\mathbf{b}_{SA}^{2}),
\end{equation}
where the meanings of the notations $\sigma \left( \cdot  \right)$, $\delta \left( \cdot  \right)$ and  $*$ are the same as those used in Eq.~(\ref{Eq: eq 2}). The first convolutional layer with parameters of $\mathbf{W}_{SA}^{1}\in {{\mathbb{R}}^{\gamma C\times C\times 1\times 1}}$ and $\mathbf{b}_{SA}^{1}\in {{\mathbb{R}}^{\gamma C}}$ is used to yield per-channel attentive maps which are then combined into a single attentive map by the second $1\times 1$ convolutional layer (parameterized by $\mathbf{W}_{SA}^{2}$ and $\mathbf{b}_{SA}^{2}$). Further, the sigmoid function $\sigma \left( \cdot  \right)$ normalizes the attentive map range to $\left[ 0,1 \right]$ to obtain the spatial attention soft mask $\bm{\upbeta}$.  The process of input features being spatially modulated by  $\bm{\upbeta}$ can be formulated as
\begin{equation}
\label{Eq: eq 5}
{{\mathbf{U}}_{SA}}={{\Phi }_{SA}}(\mathbf{U})={{f}_{SA}}(\mathbf{U}, \bm{\upbeta }),
\end{equation}
where  ${{f}_{SA}}\left( \cdot  \right)$ is an element-wise multiplication for spatial positions of each feature-map and their corresponding spatial attention weights, and ${{\Phi }_{SA}}\left( \cdot  \right)$ denotes the SA model.

With the SA unit, the features are adaptively modulated in a local way, which could be interplayed with the global channel-wise modulation to help the network enhancing the representational power.

\subsubsection{ Integration of CA and SA into the Residual Block}
Since the residual blocks introduced in ResNets \cite{19He2016CVPR_ResnetIR} can improve information flow and achieve better performance for image SR in \cite{20Lim2017CVPRW_EDSR}, we combine the channel-wise and spatial attention units into the residual block and propose the CSAR block.

As illustrated in Fig.~\ref{fig:The CSAR block}, if we denote ${{\mathbf{H}}_{I}}$ and ${{\mathbf{H}}_{O}}$ as the input and output of a CSAR block, and ${{\Phi }}\left( \cdot  \right)$ as the combinational attention model of CA and SA that will be detailed later, the CSAR block can be formulated as
\begin{equation}
\label{Eq: eq 6}
{{\mathbf{H}}_{O}}={\mathcal{Q}}({{\mathbf{H}}_{I}})={{\mathbf{H}}_{I}}+ {\Phi} \left( \mathbf{U} \right)={{\mathbf{H}}_{I}}+\Phi \left( {\mathcal{R}}({{\mathbf{H}}_{I}}) \right),
\end{equation}
where ${\mathcal{Q}}\left( \cdot  \right)$ and ${\mathcal{R}}\left( \cdot  \right)$ represent the functions of the CSAR block and the residual branch respectively. The residual branch contains two stacked convolutional layers with a ReLU activation,
\begin{equation}
\label{Eq: eq 7}
\mathbf{U}={\mathcal{R}}({{\mathbf{H}}_{I}})=\mathbf{W}_{{\mathcal{R}}}^{2}*\delta (\mathbf{W}_{{\mathcal{R}}}^{1}*{{\mathbf{H}}_{I}}+\mathbf{b}_{{\mathcal{R}}}^{1})+\mathbf{b}_{{\mathcal{R}}}^{2},
\end{equation}
where $\left\{ \mathbf{W}_{{\mathcal{R}}}^{i} \right\}_{i=1}^{2}$ and $\left\{ \mathbf{b}_{{\mathcal{R}}}^{i} \right\}_{i=1}^{2}$ are the weight and bias sets of the residual branch  and $\mathbf{U}$ is a set of produced residual features.

To capture more important information, we apply the combinational attention model  ${{\Phi }}\left( \cdot  \right)$  to modulate the residual features $\mathbf{U}$. At first, we operate the CA unit ${{\Phi }_{CA}}\left( \cdot  \right)$ and the SA unit ${{\Phi }_{SA}}\left( \cdot  \right)$ on the residual features $\mathbf{U}$  respectively to obtain channel-wise weighted feature-maps ${{\mathbf{U}}_{CA}}$ and spatial weighted feature-maps ${{\mathbf{U}}_{SA}}$,  as described in Section III. \emph{A} \emph{1)} and \emph{2)}. Then, two sets of modulated feature-maps are concatenated as the input to a $1\times 1$ convolutional layer ( parameterized by ${{\mathbf{W}}_{\Phi }}$ and ${{\mathbf{b}}_{\Phi }}$ ) which is utilized to fuse two types of attention-modulated features with learned adaptive weights. All processes are summarized as follows.
\begin{equation}
\begin{split}
\label{Eq: eq 8}
&{{\mathbf{U}}_{CA}}={{\Phi }_{CA}}\left( \mathbf{U} \right), \\
&{{\mathbf{U}}_{SA}}={{\Phi }_{SA}}\left( \mathbf{U} \right),\\
&\Phi \left( \mathbf{U} \right)={{\mathbf{W}}_{\Phi }}*\left[ {{\mathbf{U}}_{CA}},{{\mathbf{U}}_{SA}} \right]+{{\mathbf{b}}_{\Phi }},
\end {split}
\end{equation}
where  $\left[ \cdot \right]$  represents the operation of feature concatenation.

Inserting the combinational attention model into the deep network in the way described above has two benefits. First, since the combinational attention model only modulates the residual features, the good property of the identical mapping in the residual block is not broken and the information flow is still improved. Second, as two attention units are combined into a residual block, we can conveniently apply channel-wise and spatial attentions to multi-level features by stacking multiple CSAR blocks, and thus more multi-level important information is captured.

\subsection{The FMM Module}
To make full use of the attention mechanism and conveniently maintain persistent memory, the FMM module is built. As illustrated in Fig.~\ref{fig:The architectures}(b), the FMM module contains a CSAR blockchain and a gated fusion (GF) node.

The CSAR blockchain is constructed by stacking multiple CSAR blocks in a chain structure, which is exploited to perform channel-wise and spatial feature modulation at multiple levels.
Supposing $B$ CSAR blocks in a blockchain are stacked in sequence, the input of the first CSAR block ${{\mathbf{H}}^{0}}$ and the output of the last CSAR block ${{\mathbf{H}}^{B}}$ are obviously the input and output of the CSAR blockchain. Thus, the CSAR blockchain can be formulated as below.
\begin{equation}
\begin{aligned}
\label{Eq: eq 9}
{{\mathbf{H}}^{B}}={} & {\mathcal{Q}}^{(B)}({{\mathbf{H}}^{0}})\\
={} & {\mathcal{Q}}^{B}\left({\mathcal{Q}}^{B-1}\left( \cdots \left( {\mathcal{Q}}^{1}({{\mathbf{H}}^{0}}) \right)\cdots  \right) \right),
 \end{aligned}
\end{equation}
where $\left\{ {\mathcal{Q}}^{b}\left( \cdot  \right) \right\}_{b=1}^{B}$ are the functions for the CSAR blocks as depicted in Eq.~(\ref{Eq: eq 6}), and $ {\mathcal{Q}}^{(B)}(\cdot)$ denotes the operation of the CSAR blockchain.

To preserve long-term information when multiple FMM modules are stacked in the deep network, the GF node is attached to integrate the information coming from the previous FMM modules and from the current blockchain through an adaptive learning process. In the GF node, the features generated by the preceding FMM modules and by the current CSAR blockchain are firstly concatenated and then fed into a convolutional layer to be adaptively fused. Let  ${{\mathbf{P}}_{i}} (i=1,2, \cdots, m-1)$  and  $\mathbf{H}_{m}^{B}$ be the output features of $m-1$ previous  FMM modules and of  the current CSAR blockchain with $B$  CSAR blocks. The process of gated fusion is formulated as
\begin{equation}
\begin{aligned}
\label{Eq: eq 10}
{{\mathbf{P}}_{m}}={} & {{\phi }_{m}}([\mathbf{H}_{m}^{B}, {{\mathbf{P}}_{1}}, {{\mathbf{P}}_{2}}, \cdots,{{\mathbf{P}}_{m-1}}]) \\
={} & \mathbf{W}_{GF}^{m}*\left[ \mathbf{H}_{m}^{B}, {{\mathbf{P}}_{1}}, {{\mathbf{P}}_{2}}, \cdots,{{\mathbf{P}}_{m-1}} \right]+\mathbf{b}_{GF}^{m},
 \end{aligned}
\end{equation}
where ${{\phi }_{m}}\left( \cdot  \right)$  denotes the function of the $1 \times 1$  convolutional layer  with  parameters of  ${\mathbf{W}_{GF}^{m}}$ and ${\mathbf{b}_{GF}^{m}}$.  This convolutional layer  accomplishes the gating mechanism to learn adaptive weights for different information and then controls the output information. Based on those depicted above, the formulation of the $m\text{-th}$ FMM module can be written as
\begin{equation}
\begin{aligned}
\label{Eq: eq 11}
{{\mathbf{P}}_{m}}={} & {{\mathcal{G}}_{m}}({{\mathbf{P}}_{m-1}})={{\phi }_{m}}([\mathbf{H}_{m}^{B},{{\mathbf{P}}_{1}},{{\mathbf{P}}_{2}}, \cdots,{{\mathbf{P}}_{m-1}}]) \\
={} & {{\phi }_{m}}([{\mathcal{Q}}_{m}^{(B)}({{\mathbf{P}}_{m-1}}), {{\mathbf{P}}_{1}}, {{\mathbf{P}}_{2}}, \cdots,{{\mathbf{P}}_{m-1}}]),
 \end{aligned}
\end{equation}
where  ${{\mathcal{G}}_{m}}\left( \cdot  \right)$ denotes the function for the $m\text{-th}$ FMM module, and  ${{\mathbf{P}}_{m-1}}$ and ${{\mathbf{P}}_{m}}$ are the input and output of the $m\text{-th}$ FMM module.  As ${{\mathbf{P}}_{m-1}}$ is also the input of the CSAR blockchain (${\mathcal{Q}}_{m}^{(B)}(\cdot)$) in the $m\text{-th}$ FMM module (i.e., $\mathbf{H}_{m}^{0}={{\mathbf{P}}_{m-1}}$ ), there is $\mathbf{H}_{m}^{B}={\mathcal{Q}}_{m}^{(B)}({{\mathbf{P}}_{m-1}})$ in Eq.~(\ref{Eq: eq 11}).

Thus, in the CSAR blockchain, the stacked CSAR blocks modulate multi-level features to capture more important information, and  multiple short-term skip connections help rich information flow across different layers and modules. Meanwhile, in the GF node, the long-term dense connections among the FMM modules not only alleviate long-term information loss of  the deep network during forward propagation but also contribute to multi-level information fusion, which would benefit image SR.

\begin{figure}[!t]
\captionsetup[subfigure]{farskip = 0pt}
\captionsetup{belowskip=-16pt}
    \small
    \begin{minipage}[b]{1\linewidth}
        \centering
        \subfloat[BR block]{
            \centering
            \includegraphics[width=0.2\linewidth]{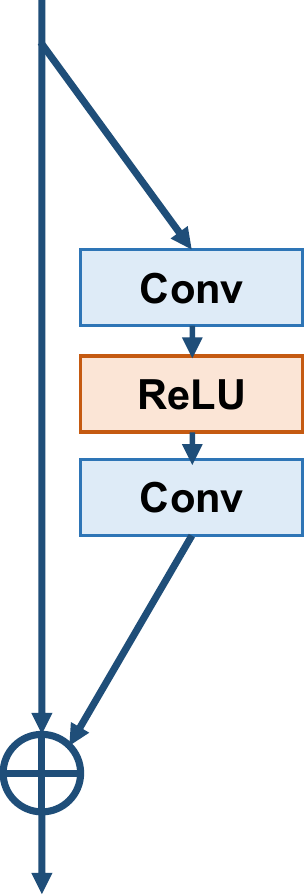}
        }
        \hspace{0.1\linewidth}
        \subfloat[CAR block]{
            \centering
            \includegraphics[width=0.2\linewidth]{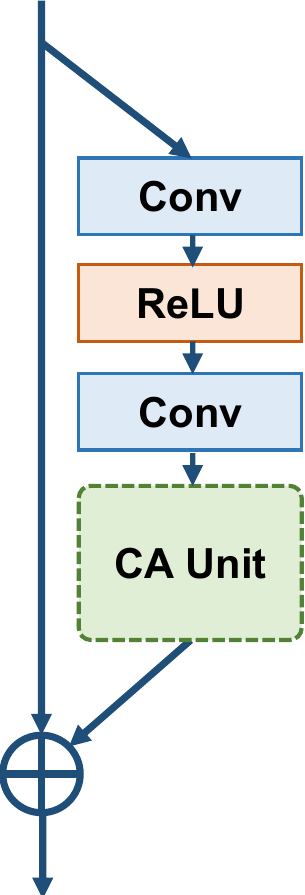}
        }
        \hspace{0.1\linewidth}
        \subfloat[SAR block]{
            \centering
            \includegraphics[width=0.2\linewidth]{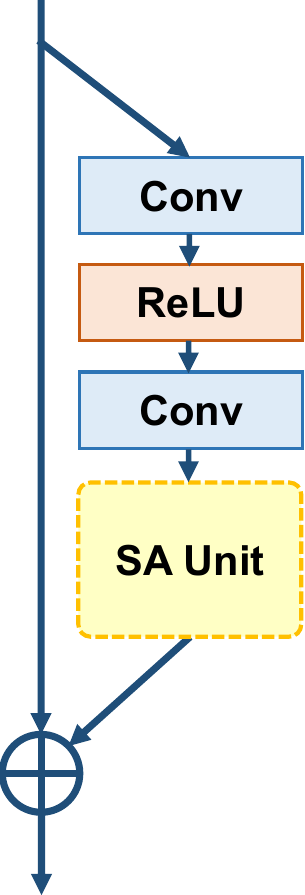}
        }
        \end{minipage}
\caption{Three other blocks for the comparisons with our CSAR block. (a) The base residual (BR) block without any form attention, which corresponds to the $\nth{1}$  and $\nth{5}$ combinations of the first three rows in TABLE~\ref{tab:Ablation study on effects of CSAR and GF}. (b) The channel-wise attention residual (CAR) block corresponding to the  $\nth{2}$  and $\nth{6}$ combinations of the first three rows in TABLE~\ref{tab:Ablation study on effects of CSAR and GF}. (c) The spatial attention residual (SAR) block corresponding to the $\nth{3}$ and $\nth{7}$ combinations of the first three rows in TABLE~\ref{tab:Ablation study on effects of CSAR and GF}.}
\label{fig:The BR CAR SAR blocks}
\end{figure}

\begin{table*}[htp]
\renewcommand\arraystretch{1.1}
\captionsetup{belowskip=-10pt,justification=centering}
\caption{\textsc{ \\ Ablation Study on Effects of the Channel-wise and Spatial Attention Residual (CSAR) Block and  \\
the Gated Fusion (GF) Node with Long-term Dense Connections. \\
Average PSNRs for a Scale Factor of $2\times$  on Urban100 Dataset Are Reported.
}}
\footnotesize
\centering
\begin{tabularx}{ 1.0\textwidth}{{|@{\extracolsep{\fill}} p{2cm}<{\centering}|c|c c c c c c c c|}}
\hline
\multicolumn{2}{|c|}{Components}& \multicolumn{8}{c|}{Different Combinations of Components}\\
\hline
\hline
\multirow{3}*{\centering{ In residual blocks}}
& Channel-wise attention (CAR) & \texttimes & \checkmark  & \texttimes & --  & \texttimes & \checkmark & \texttimes & -- \\
& Spatial attention (SAR)  &  \texttimes &  \texttimes & \checkmark & --  &  \texttimes &  \texttimes & \checkmark & -- \\
& Combinational attention of CA and SA (CSAR)  & \texttimes & --  & --  & \checkmark  & \texttimes & -- &  -- & \checkmark  \\ \cline{1-2}
\multicolumn{2}{|c|}{Gated fusion (GF) node with long-term dense connections}
& \texttimes & \texttimes & \texttimes & \texttimes & \checkmark  & \checkmark  & \checkmark  & \checkmark  \\
\hline
\multicolumn{2}{|c|}{PSNR (dB)}
& 32.38 & 32.48 & 32.44 & 32.54 & 32.48 & 32.52 & 32.50 & 32.59\\
\hline
\end{tabularx}
\label{tab:Ablation study on effects of CSAR and GF}
\end{table*}

\subsection{Network Architecture}
As shown in Fig.~\ref{fig:The architectures}, we stack multiple FMM modules to build the feature transformation sub-network (FTNet), which is utilized to map the features, generated from the initial feature extraction sub-network (IFENet), to the high informative features for the upscaling sub-network (UpNet). In addition, similar to \cite{20Lim2017CVPRW_EDSR, 24Zhang2018CVPR_RDN}, we also adopt the global residual-feature learning in the FTNet via adding an identity branch from its input to its output (green curve in  Fig.~\ref{fig:The architectures}). Thus, the three sub-networks make up our CSFM network to super-resolve LR image. Let's denote $\mathbf{X}$ and ${{\mathbf{Y}}_{SR}}$  as the input and output of the CSFM network. And, we adopt a convolutional layer as the IFENet to extract the initial features from LR input image,
\begin{equation}
\label{Eq: eq 12}
{{\mathbf{F}}_{IFE}}={{S}_{IFENet}}(\mathbf{X}),
\end{equation}
where  ${{S}_{IFENet}}(\cdot )$ denotes the function of the IFENet,  and ${{\mathbf{F}}_{IFE}}$ is a set of  extracted features which is then fed into the FTNet and also used for global residual-feature learning.

In the FTNet, the input  ${{\mathbf{F}}_{IFE}}$ is firstly sent to a convolutional layer for receptive field expansion and  the generated features ${{\mathbf{P}}_{0}}$ are then used as the input to the first FMM module. Supposing $M$ FMM modules and one convolutional layer are stacked to act as the features transformation, the output of the FTNet can be obtained by
\begin{equation}
\label{Eq: eq 13}
\begin{aligned}
{{\mathbf{F}}_{FT}}={} & {{S}_{FTNet}}({{\mathbf{F}}_{IFE}}) \\
={} &{{f}_{conv}}({{\mathcal{G}}^{(M)}}({{\mathbf{P}}_{0}}))+{{\mathbf{F}}_{IFE}}\\
={} &{{f}_{conv}}({{\mathcal{G}}_{M}}({{\mathcal{G}}_{M-1}}(\cdots ({{\mathcal{G}}_{1}}({{\mathbf{P}}_{0}}))\cdots )))+{{\mathbf{F}}_{IFE}},
\end{aligned}
\end{equation}
where  ${{S}_{FTNet}}(\cdot )$  represents the FTNet of which the output is ${{\mathbf{F}}_{FT}}$,  ${{f }_{conv}}(\cdot )$ is the convolutional operation, ${{\mathcal{G}}_{m}(\cdot)} (m=1, 2, \cdots, M)$ denotes the function for the $m\text{-th}$ FMM module as described in Eq.~(\ref{Eq: eq 11}).

After acquiring the high informative features ${{\mathbf{F}}_{FT}}$, we exploit the UpNet to upsample them for HR image reconstruction. Specifically, we adopt a sub-pixel convolutional layer \cite{17Shi2016CVPR_Sub-Pixel} followed by a convolutional layer as the UpNet for converting multiple HR sub-images to a single HR image.
\begin{equation}
\label{Eq: eq 14}
\begin{aligned}
{{\mathbf{Y}}_{SR}}={} &{D}(\mathbf{X})\\
={} & {{S}_{UpNet}}({{\mathbf{F}}_{FT}})\\
={} & {{S}_{UpNet}}({{S}_{FTNet}}({{S}_{IFENet}}(\mathbf{X}))),
\end{aligned}
\end{equation}
where ${{S}_{UpNet}}(\cdot )$ and ${D}\left( \cdot  \right)$ denote the functions of the UpNet and the whole CSFM network respectively.

The CSFM network is optimized via minimizing the difference between the super-resolved image ${{\mathbf{Y}}_{SR}}$ and the corresponding ground-truth image ${\mathbf{Y}}$. As done in previous work \cite{20Lim2017CVPRW_EDSR, 24Zhang2018CVPR_RDN}, we adopt ${{L}_{1}}$ loss function to measure the difference. Given a training dataset  $\left\{ {{\mathbf{X}}^{k}},{{\mathbf{Y}}^{k}} \right\}_{k=1}^{K}$, where $K$  is the number of training patch pairs and $\left\{ {{\mathbf{X}}^{k}},{{\mathbf{Y}}^{k}} \right\}$ are the $k\text{-th}$ LR and HR patch pairs, the objective function for training the CSFM network is formulated as
\begin{equation}
\label{Eq: eq 15}
{\mathcal{L}}(\Theta )=\frac{1}{K}\sum\limits_{k=1}^{K}{{{\left\| {{\mathbf{Y}}^{k}}-D({{\mathbf{X}}^{k}}) \right\|}_{1}}},
\end{equation}
where $\Theta $ denotes the parameter set of the CSFM network.

With the stacked FMM modules and the densely connected structure, the proposed CSFM network not only possesses the discriminative learning ability for different types of information but also enables the information that is easier to reconstruct to adopt the shorter forward/backward paths across the network and then pays  more attentions to the more important and more difficult information.

\section{Experiments and Analysis}
In this section, we first provide implementation details, including both model hyper-parameters and training data setting. Then, we study the contributions of different components in the proposed CSFM network by the ablation experiments. Finally, we compare our CSFM model with other state-of-the-art methods on several benchmark datasets.

\subsection{Datasets and Metrics}
We conduct comparison studies on widely used datasets, Set5 \cite{41Bevilacqua2012BMVC}, Set14 \cite{42Zeyde2010ICCS}, BSD100 \cite{43ArbelaezP2011TPAMI}, Urban100 \cite{44Huang2015CVPRSelfExp} and Manga109 \cite{45Matsui2017MultiTollAppli}, which contain 5, 14, 100, 100 and 109 images respectively. The Set5, Set14 and BSD100 contain natural scene images, while the Urban100 consists of urban scene images with many details in different frequency bands and Manga109 is made up of Japanese comic images with many fine structures. We use 800 high-quality training images from DIV2K \cite{46Timofte2017CVPRW_NTIRE} to train our model. Data augmentation is performed on these training images, which includes random horizontal flipping and random rotation by ${{90}^{\circ }}$.

We use the peak signal-to-noise ratio (PSNR) and the structural similarity (SSIM) \cite{47Wang2004TIP_SSIM} index as metrics for evaluation.  Higher PSNR and SSIM values indicate better quality. As commonly done in SISR, all the criteria are calculated on the luminance channel of image after pixels near image boundary are removed.

\begin{figure}[t]
\vspace*{-0.2cm}
\hspace{-0.2cm}
\captionsetup[subfloat]{labelformat=empty, belowskip=-12pt}
  \subfloat [\centerline{``{\em img015}'' from Urban100} \protect \linebreak \centerline{for $4\times$ upscaling}]
   {
        \footnotesize
        \includegraphics[width=0.49\linewidth]{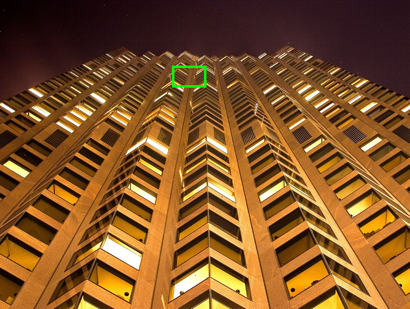}
       }
    \begin{minipage}[b]{0.78\linewidth}
        \subfloat[\centerline{(a) Ground Truth} \protect \linebreak \centerline{PSNR / SSIM}]{
            \centering
            \includegraphics[width=0.29\linewidth]{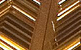}
        }
        \subfloat[\centerline{(b) BR}  \protect \linebreak \centerline{26.55 / 0.7479}]{
            \centering
            \includegraphics[width=0.29\linewidth]{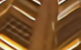}
        }\\[-0.33cm]
        \subfloat[\centerline{(c) CSAR} \protect \linebreak \centerline{26.79 / 0.7543}]{
            \centering
            \includegraphics[width=0.29\linewidth]{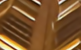}
        }
        \subfloat[\centerline{(d) CSAR$+$GF} \protect \linebreak \centerline{{\bf 26.92 / 0.7577}}]{
            \centering
            \includegraphics[width=0.29\linewidth]{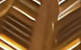}
        }
        \end{minipage}
\caption{The visual comparisons of super-resolution results by the networks with various combinations of components in TABLE~\ref{tab:Ablation study on effects of CSAR and GF}.
The assessments are made for  $4\times$ upscaling on the image ``{\em img015}'' from Urban100. (b) is the result produced by the baseline network with BR blocks corresponding to the $\nth{1}$  combination in TABLE~\ref{tab:Ablation study on effects of CSAR and GF}. (c) is generated by the network with proposed CSAR blocks corresponding to the $\nth{4}$  combination in TABLE~\ref{tab:Ablation study on effects of CSAR and GF}. (d) presents the result by the network with both the CSAR blocks and GF nodes  corresponding to the last  combination in TABLE~\ref{tab:Ablation study on effects of CSAR and GF}. It is obvious that both CSAR blocks for attentive feature-modulation and the GF nodes for long-term information maintenance contribute to generating more faithful result.}
\label{fig: Subjective comparison of BR_CSAR_CSARandGF}
\end{figure}

\subsection{Implementation Details}
We apply our model to super-resolve the RGB low-resolution images which are generated by downsampling the corresponding HR images with bicubic kernel to a certain scale. Following \cite{20Lim2017CVPRW_EDSR}, we pre-process all images by subtracting the mean RGB values of  DIV2K dataset. For training, the LR color patches with a size of $48\times48$ are randomly cropped from LR images as the inputs of our proposed model and the mini-batch size is set to 16. We train our model with ADAM optimizer \cite{48Kingma2015ICLR_Adam} by setting ${{\beta }_{1}}=0.9$, ${{\beta }_{2}}=0.999$ and $\varepsilon ={{10}^{-8}}$. The initial learning rate is initialized to ${{10}^{-4}}$, which is reduced to half  at $\text{3}\times \text{1}{{\text{0}}^{\text{5}}}$ mini-batch updates and then halved at every $\text{2}\times \text{1}{{\text{0}}^{\text{5}}}$ iterations. And, we apply PyTorch \cite{49Paszke2017NIPSWAutodiff_Pytorch} on an NVIDIA GTX 1080Ti GPU for model training and testing.

In our CSFM network, all convolutional layers have 64 filters and the kernel sizes of them are $3\times3$ except the  $1\times1$ convolutional layers in the CA and SA units and those  in the GF nodes. Meanwhile, we zero-pad the boundaries of each feature-map to ensure the spatial size of it is the same as the input size after the convolution is operated. In addition, in the CSAR block, the reduction ratio $r$ in  the CA unit and the increase ratio $\gamma $ in the SA unit are empirically set to 16 and 2 respectively.

\begin{figure}
\captionsetup[subfigure]{farskip = 0pt}
\captionsetup{belowskip=-14pt}
    \small
    \begin{minipage}[b]{1\linewidth}
        \centering
        \subfloat[On scale factor $2\times$]{
            \centering
            \includegraphics[width=0.51\linewidth]{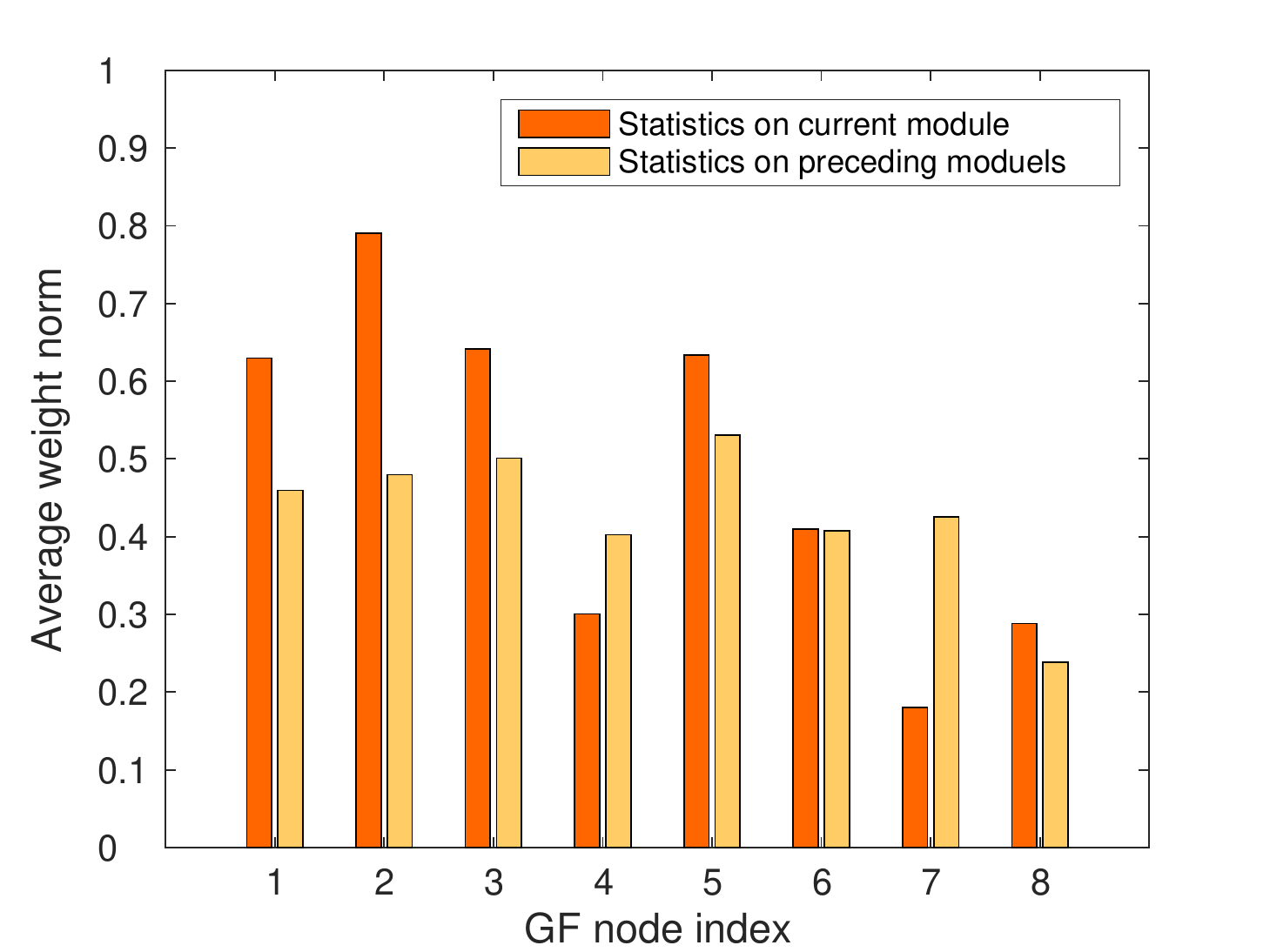}
        }
        \subfloat[On scale factor $4\times$]{
           \centering
            \includegraphics[width=0.51\linewidth]{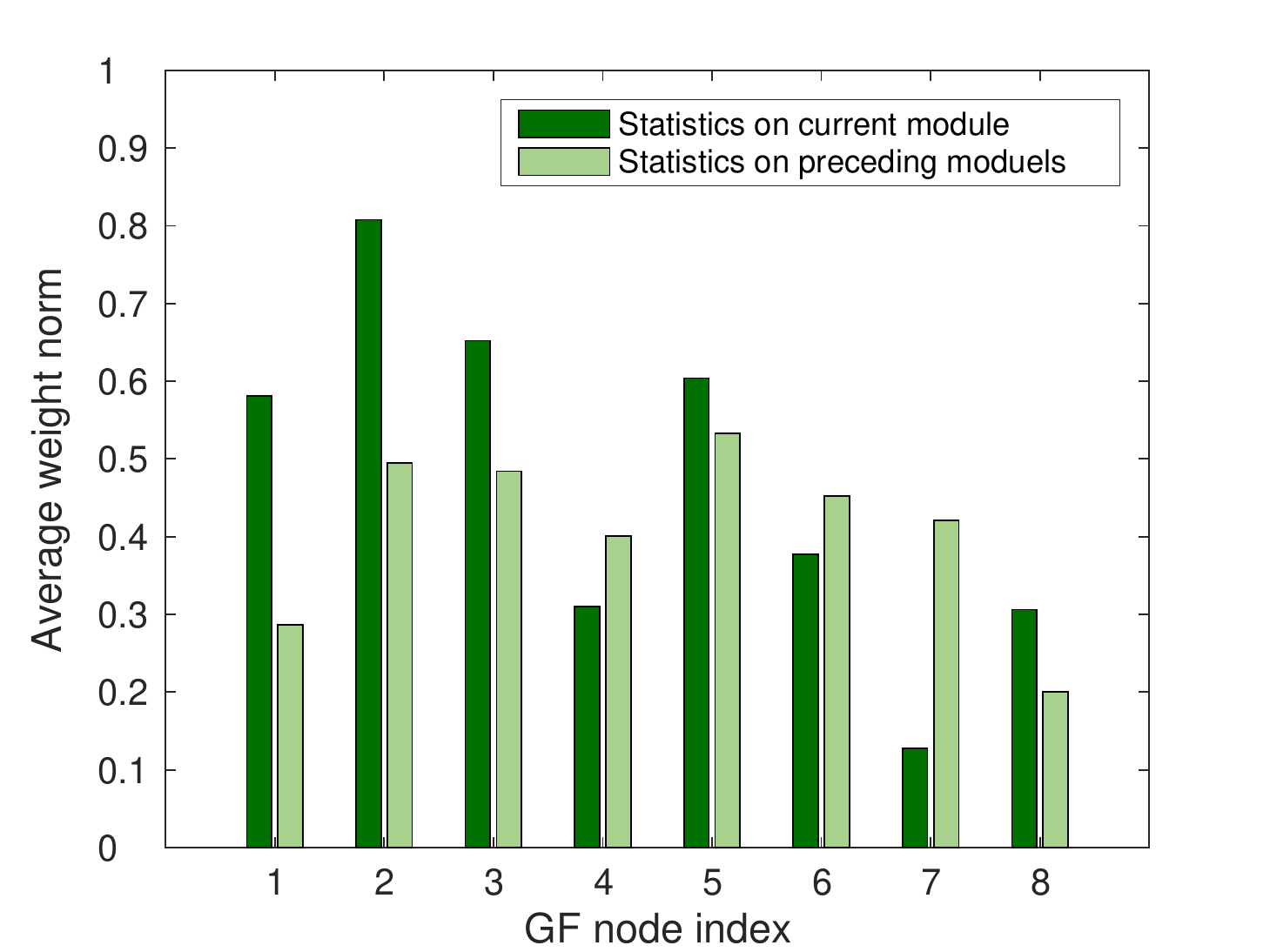}
        }
        \end{minipage}
\caption{ The average weight norms of short-term features from the current FMM module and of long-term features from the preceding FMM modules. (a) The statistics are conducted for a scale factor of $2\times$. (b) The statistics are conducted for a scale factor of $4\times$.}
\label{fig: average weight norms}
\end{figure}

\begin{figure}[t]
\captionsetup{belowskip=-18pt}
\centering
\includegraphics[width=0.9\linewidth]{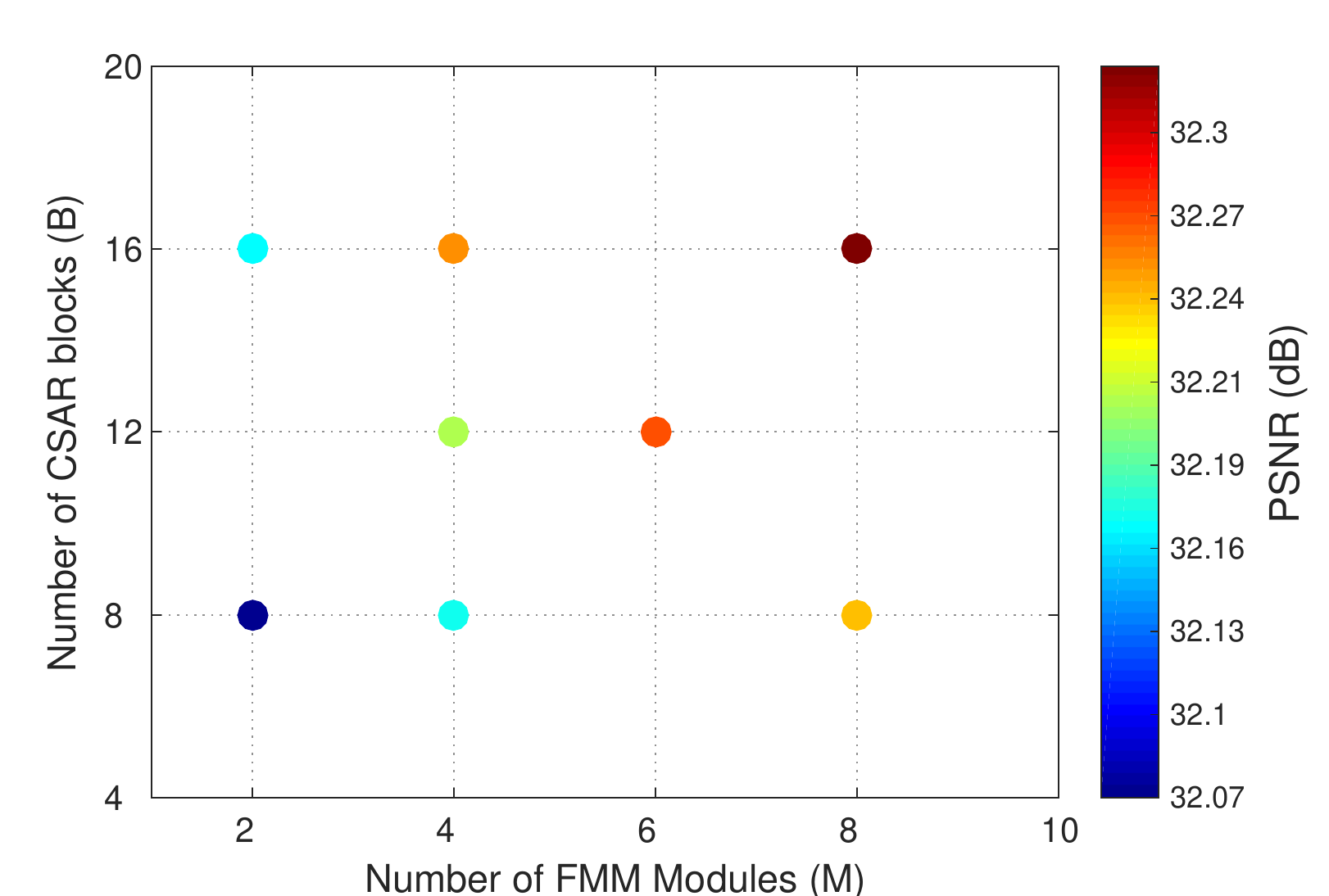}
\caption{ PSNR performance versus the number of FMM modules (M) and the number of CSAR blocks (B) per FMM . The color of the point denotes the PSNR value that corresponds to the color bar on the right. The tests are conducted for a scale factor of $2\times$ on the dataset of BSD100.}
\label{fig:The comparisons of different number modules and blocks}
\end{figure}

\subsection{Model Analysis}
In this subsection, the contributions of different components and designs in our model are analyzed via the experiments, including the CSAR block, the GF node for information persistence and the performance comparisons of different numbers of  the CSAR blocks and the FMM modules. For all experiments, all models utilized for comparisons are trained with $\text{3}\times \text{1}{{\text{0}}^{\text{5}}}$ mini-batch updates for convenience.

\subsubsection{The CSAR Block}
To validate the effectiveness of the CSAR block, besides the CSAR block, we construct another three blocks  for comparison. (I) The base residual (BR) block contains two convolutional layers with one ReLU activation, as shown in Fig.~\ref{fig:The BR CAR SAR blocks}(a). Compared with the CSAR block, the BR block  removes both the CA unit and SA unit, corresponding to the $\nth{1}$ and  $\nth{5}$ combinations of the first three rows in TABLE~\ref{tab:Ablation study on effects of CSAR and GF}. (II) The channel-wise attention residual (CAR) block is constructed by integrating the CA unit to the BR block for adaptively rescaling channel-wise features, which is depicted in Fig.~\ref{fig:The BR CAR SAR blocks}(b) and corresponds to the $\nth{2}$ and $\nth{6}$ combinations of the first three rows in TABLE~\ref{tab:Ablation study on effects of CSAR and GF}. (III) The spatial attention residual (SAR) block (the $\nth{3}$ and $\nth{7}$ combinations of the first three rows in  TABLE~\ref{tab:Ablation study on effects of CSAR and GF}), as illustrated in Fig.~\ref{fig:The BR CAR SAR blocks}(c), is developed by introducing the SA unit into the BR block to modulate pixel-wise features. Specifically, we apply 64 these blocks to the respective networks for experimental comparison, and present  SR performances of these networks on Urban100 dataset  in  TABLE~\ref{tab:Ablation study on effects of CSAR and GF}. Obviously, when the combinational attention of CA and SA is adopted in our CSAR block (the $\nth{4}$ and $\nth{8}$ combinations of the first three rows in TABLE~\ref{tab:Ablation study on effects of CSAR and GF}), the channel-wise attention or the spatial attention needs not be introduced. Therefore, we mark these cases with the symbol of  ``--''  in TABLE~\ref{tab:Ablation study on effects of CSAR and GF}.
In addition, Fig.~\ref{fig: Subjective comparison of BR_CSAR_CSARandGF}  provides the visual comparisons of the network with BR blocks (the $\nth{1}$ combination in  TABLE~\ref{tab:Ablation study on effects of CSAR and GF}), the network with CSAR blocks (the $\nth{4}$ combination in TABLE~\ref{tab:Ablation study on effects of CSAR and GF}),  and the network with both CSAR blocks and GF nodes (the last combination in TABLE~\ref{tab:Ablation study on effects of CSAR and GF}).

From TABLE~\ref{tab:Ablation study on effects of CSAR and GF}, we can see that when both the CA unit and the SA unit are removed in the BR block, the PSNR  values  are relatively low, especially when the GF nodes are not used for long-term information preservation. And, by integrating the CA unit or the SA unit into the BR blocks, the SR performances can be moderately improved. Moreover, when our proposed CSAR blocks with the combinational attentions  are utilized, the performance can be further boosted. In both cases of without and with the GF nodes, the network with the CSAR blocks outperforms those with the BR blocks by the PSNR gains of 0.16dB and 0.11dB respectively. Furthermore, in Fig.~\ref{fig: Subjective comparison of BR_CSAR_CSARandGF}, it is seen that the network only with BR blocks (Fig.~\ref{fig: Subjective comparison of BR_CSAR_CSARandGF}(b)) generates some blurry and false fence lines while the network with proposed CSAR blocks (Fig.~\ref{fig: Subjective comparison of BR_CSAR_CSARandGF}(c)) accurately reconstructs the fence rows  and presents better result via combining the channel-wise and  spatial attentions. The above observations demonstrate the superiority of our CSAR block over other blocks without attention or with only one type of attention (i.e. the BR block, CAR block and SAR block), and also manifest that integrating channel-wise and spatial attentions in residual blocks to modulate multi-level features can benefit image SR.

\begin{table*}[ht]
\captionsetup{belowskip=-10pt,justification=centering}
\caption{\textsc{ \\ Quantitative Evaluations of State-of-the-art SR Methods.\\
The Average PSNRs/SSIMs for Scale Factors of $2\times$, $3\times$ and $4\times$ Are Reported. \\
{\bf\textsc{Fontbold}} Indicates the Best Performance and {\underline{\textsc{Underline}}} Indicates the Second-best Performance.
}}
\footnotesize
\centering
\begin{tabularx}{ 0.95\textwidth}{{|@{\extracolsep{\fill}}p{1.2cm}<{\centering}|p{2.2cm}<{\centering}|c c |c c |c c |c c |c c|}}
\hline
\multirow{2}{*}{Scale} & \multirow{2}{*}{Method} & \multicolumn{2}{c|}{\textsc{Set5}} & \multicolumn{2}{c|}{\textsc{Set14}} & \multicolumn{2}{c|}{\textsc{BSD100}} & \multicolumn{2}{c|}{\textsc{Urban100}} & \multicolumn{2}{c|}{\textsc{Manga109}} \\
	& & PSNR & SSIM  & PSNR & SSIM  & PSNR & SSIM  & PSNR & SSIM & PSNR & SSIM \\
\hline
\hline
\multirow{13}*{$2\times$}
	& Bicubic
    & 33.68 & 0.9304  & 30.24 & 0.8691  & 29.56 & 0.8435  & 26.88 & 0.8405 & 30.81 & 0.9348 \\
    & SRCNN~\cite{10Dong2016TPAMI}
    & 36.66 & 0.9542 & 32.45 & 0.9067  & 31.36 & 0.8879  & 29.51 & 0.8946 & 35.70 & 0.9677 \\
    & FSRCNN~\cite{16Dong2016ECCV_FSRCNN}
    & 36.98 & 0.9556  & 32.62 & 0.9087  & 31.50 & 0.8904  & 29.85 & 0.9009 & 36.56 & 0.9703\\
    & VDSR~\cite{11KimJ2016CVPR_VDSR}
    & 37.53 & 0.9587  & 33.05 & 0.9127  & 31.90 & 0.8960  & 30.77 & 0.9141 & 37.41 & 0.9747\\
    & LapSRN~\cite{26LaiWS2017CVPR_LapSRN}
    & 37.52 & 0.9591  & 32.99 & 0.9124  & 31.80 & 0.8949  & 30.41 & 0.9101 & 37.27 & 0.9740\\
    & DRRN~\cite{12TaiY2017CVPR_DRRN}
    & 37.74 & 0.9591  & 33.23 & 0.9136  & 32.05 & 0.8973  & 31.23 & 0.9188  & 37.88 & 0.9750\\
    & MemNet~\cite{15TaiY2017ICCV_MemNet}
    & 37.78 & 0.9597  & 33.28 & 0.9142  & 32.08 & 0.8978   & 31.31 & 0.9195 & 38.02 & 0.9755\\
    & IDN~\cite{30Hui2018CVPR_IDN}
    & 37.83 & 0.9600 & 33.30 & 0.9148  & 32.08 & 0.8985 & 31.27 & 0.9196 & 38.02 & 0.9749\\
    & EDSR~\cite{20Lim2017CVPRW_EDSR}
    & 38.11 & 0.9601 & 33.92 & 0.9195  & 32.32 & 0.9013 & {\underline{32.93}} & 0.9351 & {\underline{39.19}} & {\underline{0.9782}} \\
    & SRMDNF~\cite{50Zhang2018CVPR_SRMDNF}
    & 37.79 & 0.9601 & 33.32 & 0.9154  & 32.05 & 0.8984 & 31.33 & 0.9204 & 38.07 & 0.9761 \\
    & D-DBPN~\cite{27Haris2018CVPR_DBPN}
    & 38.13 & 0.9609 & 33.83 & 0.9201  & 32.28 & 0.9009 & 32.54 & 0.9324 & 38.89 & 0.9775 \\
    & RDN~\cite{24Zhang2018CVPR_RDN}
    & {\underline{38.24}} & {\underline{0.9614}} & {\underline{34.01}} & {\underline{0.9212}}  & {\underline{32.34}} & {\underline{0.9017}} & 32.89 & {\underline{0.9353}} & 39.18 & 0.9780 \\
    & CSFM (ours)
    & {\bf 38.26} & {\bf 0.9615}  & {\bf 34.07} & {\bf 0.9213}  & {\bf 32.37} & {\bf 0.9021}  & {\bf 33.12} & {\bf 0.9366} & {\bf39.40} & {\bf0.9785}\\
\hline
\multirow{12}*{$3\times$}
    & Bicubic
    & 30.40 & 0.8686 & 27.54 & 0.7741  & 27.21 & 0.7389  & 24.46 & 0.7349 & 26.95 & 0.8565 \\
    & SRCNN~\cite{10Dong2016TPAMI}
    & 32.75 & 0.9090  & 29.29 & 0.8215  & 28.41 & 0.7863  & 26.24 & 0.7991 & 30.56 & 0.9125 \\
    & FSRCNN~\cite{16Dong2016ECCV_FSRCNN}
    & 33.16 & 0.9140  & 29.42 & 0.8242  & 28.52 & 0.7893  & 26.41 & 0.8064 & 31.12 & 0.9196  \\
    & VDSR~\cite{11KimJ2016CVPR_VDSR}
    & 33.66 & 0.9213  & 29.78 & 0.8318  & 28.83 & 0.7976  & 27.14 & 0.8279 & 32.13 & 0.9348 \\
    & LapSRN~\cite{26LaiWS2017CVPR_LapSRN}
    & 33.82 & 0.9227  & 29.79 & 0.8320  & 28.82 & 0.7973 & 27.07 & 0.8271 & 32.21 & 0.9344 \\
    & DRRN~\cite{12TaiY2017CVPR_DRRN}
    & 34.03 & 0.9244  & 29.96 & 0.8349  & 28.95 & 0.8004  & 27.53 & 0.8378 & 32.74 & 0.9388\\
    & MemNet~\cite{15TaiY2017ICCV_MemNet}
    & 34.09 & 0.9248  & 30.00 & 0.8350  & 28.96 & 0.8001  & 27.56 & 0.8376 & 32.79 & 0.9391\\
    & IDN~\cite{30Hui2018CVPR_IDN}
    & 34.11 & 0.9253  & 29.99 & 0.8354  & 28.95 & 0.8013 & 27.42 & 0.8359 & 32.69 & 0.9378 \\
    & EDSR~\cite{20Lim2017CVPRW_EDSR}
    & 34.65 & 0.9282  & 30.52 & 0.8462  & 29.25 & {\underline{0.8093}} & {\underline{28.80}} & {\underline{0.8653}} & {\underline{34.20}} & {\underline{0.9486}} \\
    & SRMDNF~\cite{50Zhang2018CVPR_SRMDNF}
    & 34.12 & 0.9254  & 30.04 & 0.8371  & 28.97 & 0.8025 & 27.57 & 0.8398 & 33.00 & 0.9403 \\
    & RDN~\cite{24Zhang2018CVPR_RDN}
    & {\underline{34.71}} & {\underline{0.9296}}  & {\underline{30.57}} &{\underline{0.8468}}  & {\underline{29.26}} & {\underline{0.8093}} & {\underline{28.80}} & {\underline{0.8653}} & 34.13 & 0.9484 \\
    & CSFM (ours)
    & {\bf 34.76} & {\bf 0.9301}  & {\bf 30.63} & {\bf  0.8477}  & {\bf  29.30} & {\bf  0.8105}  & {\bf  28.98} & {\bf  0.8681} & {\bf  34.52} & {\bf  0.9502}\\
\hline
\multirow{13}*{$4\times$}
	& Bicubic
    & 28.43 & 0.8109  & 26.00 & 0.7023  & 25.96 & 0.6678  & 23.14 & 0.6574 & 24.89 & 0.7875 \\
    & SRCNN~\cite{10Dong2016TPAMI}
    & 30.48 & 0.8628  & 27.50 & 0.7513  & 26.90 & 0.7103  & 24.52 & 0.7226 & 27.63 & 0.8553 \\
    & FSRCNN~\cite{16Dong2016ECCV_FSRCNN}
    & 30.70 & 0.8657  & 27.59 & 0.7535  & 26.96 & 0.7128  & 24.60 & 0.7258  & 27.85 & 0.8557 \\
    & VDSR~\cite{11KimJ2016CVPR_VDSR}
    & 31.35 & 0.8838  & 28.02 & 0.7678  & 27.29 & 0.7252  & 25.18 & 0.7525 & 28.87 & 0.8865 \\
    & LapSRN~\cite{26LaiWS2017CVPR_LapSRN}
    & 31.54 & 0.8866  & 28.09 & 0.7694  & 27.32 & 0.7264  & 25.21 & 0.7553 & 29.09 & 0.8893 \\
    & DRRN~\cite{12TaiY2017CVPR_DRRN}
    & 31.68 & 0.8888  & 28.21 & 0.7720  & 27.38 & 0.7284  & 25.44 & 0.7638  & 29.45 & 0.8946 \\
    & MemNet~\cite{15TaiY2017ICCV_MemNet}
    & 31.74 & 0.8893  & 28.26 & 0.7723  & 27.40 & 0.7281  & 25.50 & 0.7630  & 29.64 & 0.8971 \\
    & IDN~\cite{30Hui2018CVPR_IDN}
    & 31.82 & 0.8903  & 28.25 & 0.7730  & 27.41 & 0.7297  & 25.41 & 0.7632 & 29.41 & 0.8936 \\
    & EDSR~\cite{20Lim2017CVPRW_EDSR}
    & 32.46 & 0.8968  & 28.80 & {\underline{0.7876}}  & 27.71 & {\underline{0.7420}}  & {\underline{26.64}} & {\underline{0.8033}} & {\underline{31.03}} & {\underline{0.9158}} \\
    & SRMDNF~\cite{50Zhang2018CVPR_SRMDNF}
    & 31.96 & 0.8925  & 28.35 & 0.7772  & 27.49 & 0.7337  & 25.68 & 0.7731 & 30.09 & 0.9024 \\
    & D-DBPN~\cite{27Haris2018CVPR_DBPN}
    & 32.42 & 0.8977  & 28.76 & 0.7862  & 27.68 & 0.7393  & 26.38 & 0.7946 & 30.91 & 0.9137 \\
    & RDN~\cite{24Zhang2018CVPR_RDN}
    & {\underline{32.47}} & {\underline{0.8990}}  & {\underline{28.81}} & 0.7871  & {\underline{27.72}} & 0.7419  & 26.61 & 0.8028 & 31.00 & 0.9151 \\
    & CSFM (ours)
    & {\bf 32.61} & {\bf 0.9000} & {\bf 28.87} & {\bf 0.7886}  & {\bf 27.76} & {\bf 0.7432} & {\bf 26.78} & {\bf 0.8065} & {\bf 31.32} & {\bf 0.9183}\\
\hline
\end{tabularx}
\label{tab:Quantitative evaluations of state-of-the-art SR methods}
\end{table*}

\subsubsection{The GF Node with  Long-term Dense Connections}
As illustrated in Fig.~\ref{fig:The architectures}(b), the GF node is added at the end of the FMM module for contributing to persistent memory maintenance and different information fusion. To investigate the contributions of the GF node, we conduct the ablation tests and present the study on the effect of the GF node  in TABLE~\ref{tab:Ablation study on effects of CSAR and GF} and Fig.~\ref{fig: Subjective comparison of BR_CSAR_CSARandGF}. In TABLE~\ref{tab:Ablation study on effects of CSAR and GF},  the first four columns list the results produced by the networks without  GF nodes where 64 blocks are cascaded for feature transformation, while the last four columns show the performances of the networks with  GF nodes in which 16 blocks and one GF node constitute a module and 4  modules are stacked with densely connected structure  (similar to the architecture of the CSFM network). Through the comparisons between the results in the first four columns and those in the last four columns, we find that the networks with  GF nodes would perform better than those without  GF nodes. Specifically, when the CSAR blocks with combinational attentions are utilized, the network with  GF nodes can achieve an improvement of 0.21dB in terms of PSNR compared with the baseline network with only BR blocks. Besides, from Fig.~\ref{fig: Subjective comparison of BR_CSAR_CSARandGF}, we can observe that by introducing information maintenance mechanism, the network with GF nodes  generates finer and clearer fence rows compared with those without GF nodes.  These comparisons manifest that applying the GF nodes makes long-term information preservation easy and then more important information can be effectively exploited for image SR.

To further analyze the contributions of different kinds of information fed into the GF nodes and illustrate how the GF nodes control the output information, and inspired by \cite{15TaiY2017ICCV_MemNet}, we make statistics on the norms of the weights from all filters in the GF nodes. For each feature-map input to the GF node, we calculate the weight norm in the corresponding filter as follows
\begin{equation}
\label{Eq: eq 16}
q_{n}^{m}=\sqrt{\sum\limits_{i=1}^{64}{{{\left( \mathbf{W}_{GF}^{m}\left[ i,n,1,1 \right] \right)}^{2}}}}
\end{equation}
where  $q_{n}^{m} \left( n=1,2,\cdots, {{N}_{m}} \right)$ represents the weight norm of the $n\text{-th}$ feature-map fed into the $m\text{-th}$ GF node (receiving ${{N}_{m}}$  feature-maps as input), and $\mathbf{W}_{GF}^{m}$ with size of $64\times {{N}_{m}}\times 1\times 1$ denotes the weight set of the filter in the GF node. The larger norm indicates that the feature-map provides more information to the GF node for fusion, and vice versa. For the sake of comparison, we average the weight norms of long-term feature-maps from the preceding FMM modules and of short-term feature-maps from the current FMM module respectively. Similar to \cite{15TaiY2017ICCV_MemNet}, we normalize the weight norms to the range of 0 to 1 for better visualization. Fig.~\ref{fig: average weight norms} presents the average norms of two types of feature-maps (long-term feature-maps and short-term feature-maps) in eight GF nodes of eight FMM modules for two scale factors of  $2\times$  and $4\times$. One can see that the long-term information from the  preceding modules makes non-negligible contribution especially in late modules whatever the upscaling factor is, which indicates that the long-term information plays an important role in super-resolving LR image. Therefore, the GF nodes being added for information persistence is beneficial for improving SR performance.

\begin{figure*}[htp]
\centering
\captionsetup{belowskip=2pt,aboveskip=3pt}
\footnotesize
   \begin{tabular}{cccccc}
   	\centering
  \hspace{-0.42cm}
    \begin{adjustbox}{valign=t}
    \begin{tabular}{c}
    \includegraphics[width=0.162\linewidth]{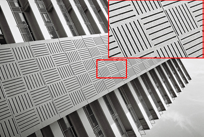}
    \\ 
     Ground Truth
     \\
     PSNR / SSIM
     \\
      \includegraphics[width=0.162\linewidth]{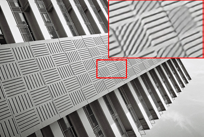}
     \\
     IDN \cite{30Hui2018CVPR_IDN}
     \\
     18.27 / 0.6176
    \end{tabular}
    \end{adjustbox}
    \hspace{-0.55cm}
    \begin{adjustbox}{valign=t}
    \begin{tabular}{c}
    \includegraphics[width=0.162\linewidth]{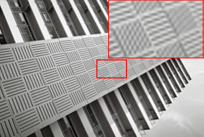}
    \\
    Bicubic
    \\
    16.58 / 0.4374
    \\
     \includegraphics[width=0.162\linewidth]{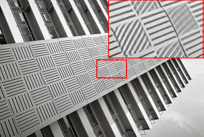}
    \\
    EDSR \cite{20Lim2017CVPRW_EDSR}
    \\
    19.14 / 0.6779
    \end{tabular}
    \end{adjustbox}
    \hspace{-0.55cm}
    \begin{adjustbox}{valign=t}
    \begin{tabular}{c}
    \includegraphics[width=0.162\linewidth]{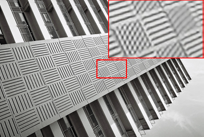} %
    \\
    SRCNN \cite{10Dong2016TPAMI}
    \\
    17.56 / 0.5413
    \\
     \includegraphics[width=0.162\linewidth]{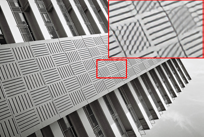} %
    \\
    SRMDNF \cite{50Zhang2018CVPR_SRMDNF}
    \\
    18.57 / 0.6308
    \end{tabular}
    \end{adjustbox}
    \hspace{-0.55cm}
    \begin{adjustbox}{valign=t}
    \begin{tabular}{c}
    \includegraphics[width=0.162\linewidth]{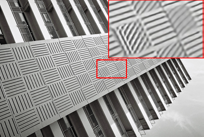} %
    \\
    VDSR \cite{11KimJ2016CVPR_VDSR}
    \\
    18.14 / 0.6011
    \\
     \includegraphics[width=0.162\linewidth]{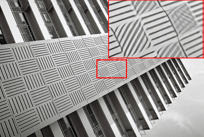} %
    \\
    D-DBPN \cite{27Haris2018CVPR_DBPN}
    \\
    18.92 / 0.6602
    \end{tabular}
    \end{adjustbox}
    \hspace{-0.55cm}
    \begin{adjustbox}{valign=t}
    \begin{tabular}{c}
    \includegraphics[width=0.162\linewidth]{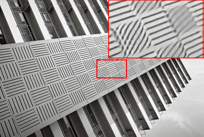} %
    \\
    LapSRN \cite{26LaiWS2017CVPR_LapSRN}
    \\
    18.20 / 0.6078
    \\
     \includegraphics[width=0.162\linewidth]{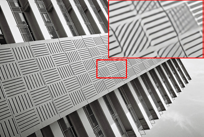}
    \\
    RDN \cite{24Zhang2018CVPR_RDN}
    \\
    19.18 / 0.6770
    \end{tabular}
    \end{adjustbox}
    \hspace{-0.55cm}
    \begin{adjustbox}{valign=t}
    \begin{tabular}{c}
    \includegraphics[width=0.162\linewidth]{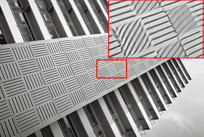} 
    \\
    MemNet \cite{15TaiY2017ICCV_MemNet}
    \\
    18.59 / 0.6397
    \\
     \includegraphics[width=0.162\linewidth]{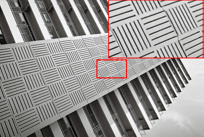}
    \\
    CSFM (Ours)
    \\
   {\bf {20.17} / \bf{0.7157}}
    \end{tabular}
    \end{adjustbox}
   \end{tabular}
\caption{Visual evaluation for a scale factor of $4\times$ on the image ``{\em img092}'' from Urban100. Our CSFM network accurately reconstructs clearer stripes while  other methods produce  blurry results with wrong directions.}
\label{fig: img092_4x}
\end{figure*}


\begin{figure*}[htp]
\centering
\captionsetup{belowskip=2pt,aboveskip=3pt}
\footnotesize
   \begin{tabular}{cccccc}
   	\centering
  \hspace{-0.42cm}
    \begin{adjustbox}{valign=t}
    \begin{tabular}{c}
    \includegraphics[width=0.162\linewidth]{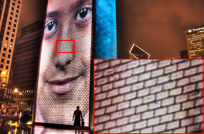}
    \\ 
     Ground Truth
     \\
     PSNR / SSIM
     \\
      \includegraphics[width=0.162\linewidth]{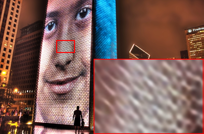} %
     \\
     IDN \cite{30Hui2018CVPR_IDN}
     \\
     22.22 / 0.6974
    \end{tabular}
    \end{adjustbox}
    \hspace{-0.55cm}
    \begin{adjustbox}{valign=t}
    \begin{tabular}{c}
    \includegraphics[width=0.162\linewidth]{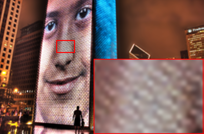}
    \\
    Bicubic
    \\
    21.57 / 0.6287
    \\
     \includegraphics[width=0.162\linewidth]{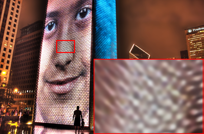}
    \\
    EDSR \cite{20Lim2017CVPRW_EDSR}
    \\
    23.94 / 0.7746
    \end{tabular}
    \end{adjustbox}
    \hspace{-0.55cm}
    \begin{adjustbox}{valign=t}
    \begin{tabular}{c}
    \includegraphics[width=0.162\linewidth]{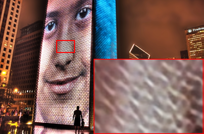} %
    \\
    FSRCNN \cite{16Dong2016ECCV_FSRCNN}
    \\
    22.00 / 0.6769
    \\
     \includegraphics[width=0.162\linewidth]{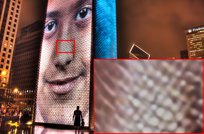} %
    \\
    SRMDNF \cite{50Zhang2018CVPR_SRMDNF}
    \\
    22.46 / 0.7109
    \end{tabular}
    \end{adjustbox}
    \hspace{-0.55cm}
    \begin{adjustbox}{valign=t}
    \begin{tabular}{c}
    \includegraphics[width=0.162\linewidth]{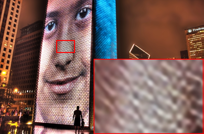} %
    \\
    VDSR \cite{11KimJ2016CVPR_VDSR}
    \\
    22.15 / 0.6920
    \\
     \includegraphics[width=0.162\linewidth]{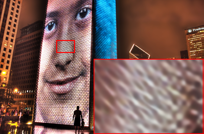}  %
    \\
    D-DBPN \cite{27Haris2018CVPR_DBPN}
    \\
    23.19 / 0.7439
    \end{tabular}
    \end{adjustbox}
    \hspace{-0.55cm}
    \begin{adjustbox}{valign=t}
    \begin{tabular}{c}
    \includegraphics[width=0.162\linewidth]{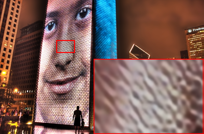} %
    \\
    LapSRN \cite{26LaiWS2017CVPR_LapSRN}
    \\
    22.01 / 0.6917
    \\
     \includegraphics[width=0.162\linewidth]{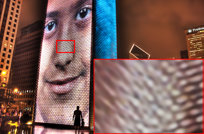}
    \\
    RDN \cite{24Zhang2018CVPR_RDN}
    \\
    24.07 / 0.7799
    \end{tabular}
    \end{adjustbox}
    \hspace{-0.55cm}
    \begin{adjustbox}{valign=t}
    \begin{tabular}{c}
    \includegraphics[width=0.162\linewidth]{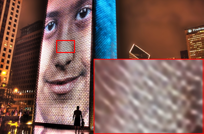}  %
    \\
    DRRN \cite{12TaiY2017CVPR_DRRN}
    \\
    21.93 / 0.6897
    \\
     \includegraphics[width=0.162\linewidth]{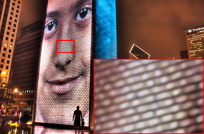}
    \\
    CSFM (Ours)
    \\
  {\bf{24.31} / \bf{0.7858}}
    \end{tabular}
    \end{adjustbox}
   \end{tabular}
\caption{Visual evaluation for a scale factor of $4\times$ on the image ``{\em img076}'' from Urban100. Other methods fail to recover the texture region on the face and give very tangle results. By contrast, our CSFM model can reconstruct the details which are subjectively closer to the ground truth.}
\label{fig: img076_x4}
\end{figure*}


\begin{figure*}[htp]
\centering
\captionsetup{belowskip=5pt,aboveskip=3pt}
\footnotesize
   \begin{tabular}{cccccc}
   	\centering
  \hspace{-0.42cm}
    \begin{adjustbox}{valign=t}
    \begin{tabular}{c}
    \includegraphics[width=0.162\linewidth]{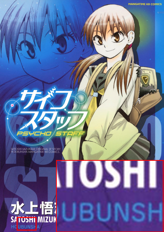}
    \\ 
    Ground Truth
     \\
     PSNR / SSIM
     \\
      \includegraphics[width=0.162\linewidth]{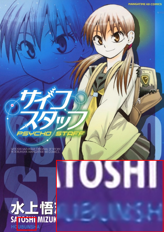} %
     \\
     IDN \cite{30Hui2018CVPR_IDN}
     \\
     33.13 / 0.9428
    \end{tabular}
    \end{adjustbox}
    \hspace{-0.55cm}
    \begin{adjustbox}{valign=t}
    \begin{tabular}{c}
    \includegraphics[width=0.162\linewidth]{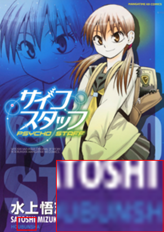}
    \\
    Bicubic
    \\
    27.52 / 0.8578
    \\
     \includegraphics[width=0.162\linewidth]{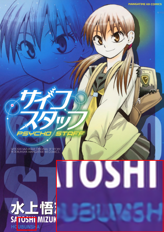}
    \\
    EDSR \cite{20Lim2017CVPRW_EDSR}
    \\
    34.60 / 0.9559
    \end{tabular}
    \end{adjustbox}
    \hspace{-0.55cm}
    \begin{adjustbox}{valign=t}
    \begin{tabular}{c}
    \includegraphics[width=0.162\linewidth]{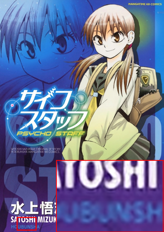} 
    \\
    SRCNN \cite{10Dong2016TPAMI}
    \\
    30.84 / 0.9123
    \\
     \includegraphics[width=0.162\linewidth]{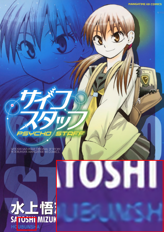}%
    \\
    SRMDNF \cite{50Zhang2018CVPR_SRMDNF}
    \\
    33.52 / 0.9476
    \end{tabular}
    \end{adjustbox}
    \hspace{-0.55cm}
    \begin{adjustbox}{valign=t}
    \begin{tabular}{c}
    \includegraphics[width=0.162\linewidth]{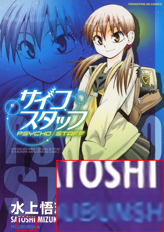} %
    \\
    VDSR \cite{11KimJ2016CVPR_VDSR}
    \\
   32.26 / 0.9387
    \\
     \includegraphics[width=0.162\linewidth]{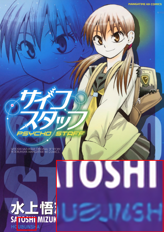} %
    \\
    D-DBPN \cite{27Haris2018CVPR_DBPN}
    \\
   34.39 / 0.9545
    \end{tabular}
    \end{adjustbox}
    \hspace{-0.55cm}
    \begin{adjustbox}{valign=t}
    \begin{tabular}{c}
    \includegraphics[width=0.162\linewidth]{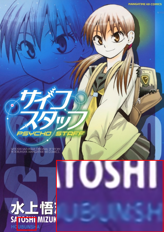} %
    \\
    LapSRN \cite{26LaiWS2017CVPR_LapSRN}
    \\
   32.57 / 0.9396
    \\
     \includegraphics[width=0.162\linewidth]{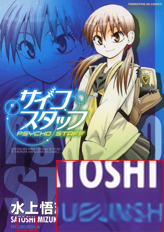}
    \\
    RDN \cite{24Zhang2018CVPR_RDN}
    \\
    34.41 / 0.9558
    \end{tabular}
    \end{adjustbox}
    \hspace{-0.55cm}
    \begin{adjustbox}{valign=t}
    \begin{tabular}{c}
    \includegraphics[width=0.162\linewidth]{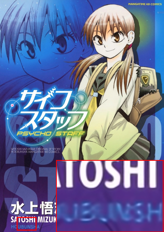}
    \\
    MemNet \cite{15TaiY2017ICCV_MemNet}
    \\
    32.99 / 0.9445
    \\
     \includegraphics[width=0.162\linewidth]{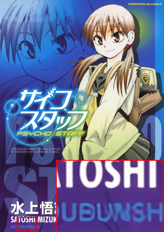}
    \\
    CSFM (Ours)
    \\
   {\bf {35.01} / \bf{0.9586}}
    \end{tabular}
    \end{adjustbox}
   \end{tabular}
\caption{Visual evaluation for a scale factor of $4\times$ on the image ``{\em PsychoStaff}'' from Manga109. Only our CSFM network can recover more recognizable characters which are too vague to be recognized in other results.}
\label{fig: PsychoStaff_x4}
\end{figure*}


\begin{figure*}[htp]
\centering
\captionsetup{belowskip=5pt,aboveskip=3pt}
\footnotesize
   \begin{tabular}{cccccc}
   	\centering
  \hspace{-0.42cm}
    \begin{adjustbox}{valign=t}
    \begin{tabular}{c}
    \includegraphics[width=0.162\linewidth]{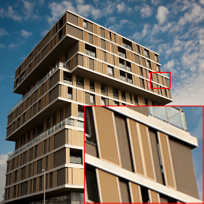}
    \\ 
     Ground Truth
     \\
     PSNR / SSIM
     \\
      \includegraphics[width=0.162\linewidth]{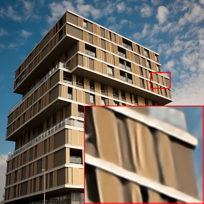}  %
     \\
     IDN \cite{30Hui2018CVPR_IDN}
     \\
     26.66 / 0.8600
    \end{tabular}
    \end{adjustbox}
    \hspace{-0.55cm}
    \begin{adjustbox}{valign=t}
    \begin{tabular}{c}
    \includegraphics[width=0.162\linewidth]{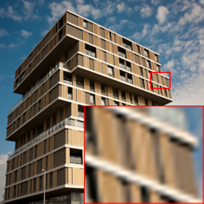}
    \\
    Bicubic
    \\
    22.86 / 0.7403
    \\
     \includegraphics[width=0.162\linewidth]{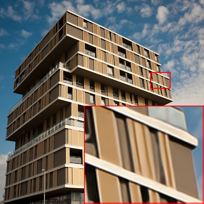}
    \\
    EDSR \cite{20Lim2017CVPRW_EDSR}
    \\
    28.97 / 0.9076
    \end{tabular}
    \end{adjustbox}
    \hspace{-0.55cm}
    \begin{adjustbox}{valign=t}
    \begin{tabular}{c}
    \includegraphics[width=0.162\linewidth]{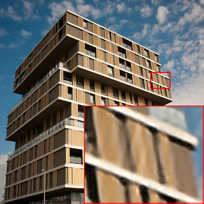} %
    \\
    FSRCNN \cite{16Dong2016ECCV_FSRCNN}
    \\
   24.84 / 0.8034
    \\
     \includegraphics[width=0.162\linewidth]{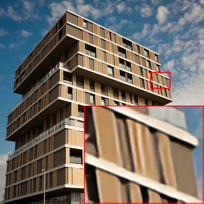} %
    \\
    SRMDNF \cite{50Zhang2018CVPR_SRMDNF}
    \\
    27.19 / 0.8744
    \end{tabular}
    \end{adjustbox}
    \hspace{-0.55cm}
    \begin{adjustbox}{valign=t}
    \begin{tabular}{c}
    \includegraphics[width=0.162\linewidth]{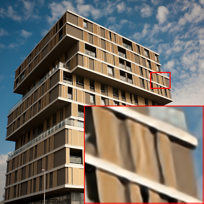} %
    \\
    VDSR \cite{11KimJ2016CVPR_VDSR}
    \\
    25.92 / 0.8428
    \\
     \includegraphics[width=0.162\linewidth]{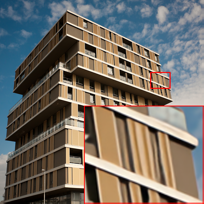}%
    \\
    D-DBPN \cite{27Haris2018CVPR_DBPN}
    \\
    28.22 / 0.8937
    \end{tabular}
    \end{adjustbox}
    \hspace{-0.55cm}
    \begin{adjustbox}{valign=t}
    \begin{tabular}{c}
    \includegraphics[width=0.162\linewidth]{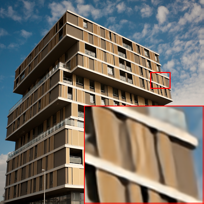} %
    \\
    LapSRN \cite{26LaiWS2017CVPR_LapSRN}
    \\
    25.98 / 0.8455
    \\
     \includegraphics[width=0.162\linewidth]{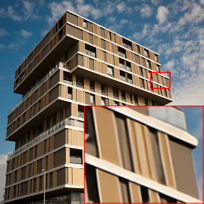}
    \\
    RDN \cite{24Zhang2018CVPR_RDN}
    \\
   28.94 / 0.9072
    \end{tabular}
    \end{adjustbox}
    \hspace{-0.55cm}
    \begin{adjustbox}{valign=t}
    \begin{tabular}{c}
    \includegraphics[width=0.162\linewidth]{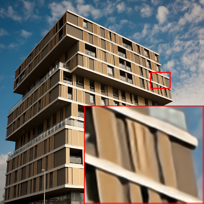}%
    \\
    DRRN \cite{12TaiY2017CVPR_DRRN}
    \\
    26.57 / 0.8608
    \\
     \includegraphics[width=0.162\linewidth]{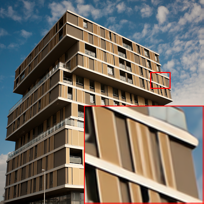}
    \\
    CSFM (Ours)
    \\
  {\bf {29.45} / \bf{0.9138}}
    \end{tabular}
    \end{adjustbox}
   \end{tabular}
\caption{Visual evaluation for a scale factor of $4\times$ on the image ``{\em img087}'' from Urban100. Only our CSFM model correctly reconstructs the color lines on the balcony while other methods generate fuzzier lines with wrong colors and structures.}
\label{fig: img087_x4}
\end{figure*}


\begin{figure*}[!htbp]
\centering
\captionsetup{belowskip=-8pt, aboveskip=3pt}
\vspace{-0.2cm}
\footnotesize
   \begin{tabular}{cccccc}
   	\centering
  \hspace{-0.42cm}
    \begin{adjustbox}{valign=t}
    \begin{tabular}{c}
    \includegraphics[width=0.162\linewidth]{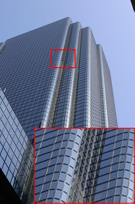}
    \\ 
     Ground Truth
     \\
     PSNR / SSIM
     \\
      \includegraphics[width=0.162\linewidth]{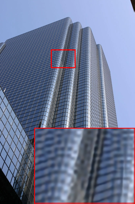} %
     \\
     IDN \cite{30Hui2018CVPR_IDN}
     \\
     23.01 / 0.6591
    \end{tabular}
    \end{adjustbox}
    \hspace{-0.55cm}
    \begin{adjustbox}{valign=t}
    \begin{tabular}{c}
    \includegraphics[width=0.162\linewidth]{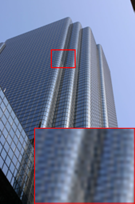}
    \\
    Bicubic
    \\
    22.16 / 0.5552
    \\
     \includegraphics[width=0.162\linewidth]{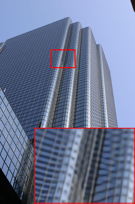}
    \\
    EDSR \cite{20Lim2017CVPRW_EDSR}
    \\
    24.22 / 0.7351
    \end{tabular}
    \end{adjustbox}
    \hspace{-0.55cm}
    \begin{adjustbox}{valign=t}
    \begin{tabular}{c}
    \includegraphics[width=0.162\linewidth]{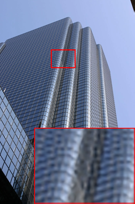} %
    \\
    SRCNN \cite{10Dong2016TPAMI}
    \\
    22.73 / 0.6133
    \\
     \includegraphics[width=0.162\linewidth]{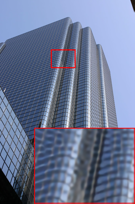} %
    \\
    SRMDNF \cite{50Zhang2018CVPR_SRMDNF}
    \\
    23.41 / 0.6753
    \end{tabular}
    \end{adjustbox}
    \hspace{-0.55cm}
    \begin{adjustbox}{valign=t}
    \begin{tabular}{c}
    \includegraphics[width=0.162\linewidth]{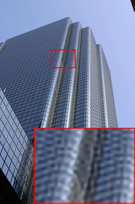}%
    \\
    VDSR \cite{11KimJ2016CVPR_VDSR}
    \\
    23.09 / 0.6415
    \\
     \includegraphics[width=0.162\linewidth]{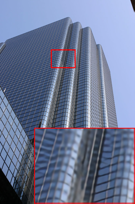}%
    \\
    D-DBPN \cite{27Haris2018CVPR_DBPN}
    \\
    23.99 / 0.7232
    \end{tabular}
    \end{adjustbox}
    \hspace{-0.55cm}
    \begin{adjustbox}{valign=t}
    \begin{tabular}{c}
    \includegraphics[width=0.162\linewidth]{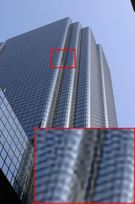}%
    \\
    LapSRN \cite{26LaiWS2017CVPR_LapSRN}
    \\
    23.15 / 0.6520
    \\
     \includegraphics[width=0.162\linewidth]{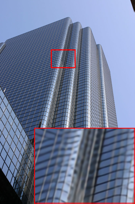}
    \\
    RDN \cite{24Zhang2018CVPR_RDN}
    \\
    24.29 / 0.7445
    \end{tabular}
    \end{adjustbox}
    \hspace{-0.55cm}
    \begin{adjustbox}{valign=t}
    \begin{tabular}{c}
    \includegraphics[width=0.162\linewidth]{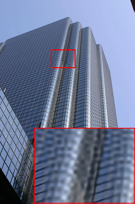} %
    \\
    MemNet \cite{15TaiY2017ICCV_MemNet}
    \\
    23.22 / 0.6676
    \\
     \includegraphics[width=0.162\linewidth]{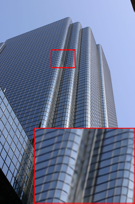}
    \\
    CSFM (Ours)
    \\
  {\bf{24.38} / \bf{0.7544}}
    \end{tabular}
    \end{adjustbox}
   \end{tabular}
\caption{Visual evaluation for a scale factor of $4\times$ on the image ``{\em img074}'' from Urban100. The reconstructed grids produced by our CSFM network are more faithful and sharper  than those by other methods.}
\label{fig: img074_x4}
\end{figure*}


\begin{figure*}[htbp]
\centering
\captionsetup{belowskip=-8pt,aboveskip=3pt}
\footnotesize
   \begin{tabular}{cccccc}
   	\centering
  \hspace{-0.42cm}
    \begin{adjustbox}{valign=t}
    \begin{tabular}{c}
    \includegraphics[width=0.162\linewidth]{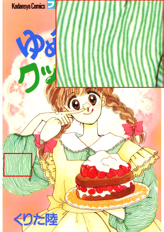}
    \\ 
     Ground Truth
     \\
     PSNR / SSIM
     \\
      \includegraphics[width=0.162\linewidth]{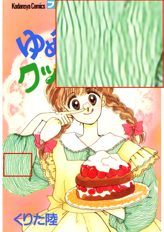}  %
     \\
     IDN \cite{30Hui2018CVPR_IDN}
     \\
     27.36 / 0.8879
    \end{tabular}
    \end{adjustbox}
    \hspace{-0.55cm}
    \begin{adjustbox}{valign=t}
    \begin{tabular}{c}
    \includegraphics[width=0.162\linewidth]{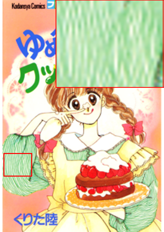}
    \\
    Bicubic
    \\
    24.66 / 0.7861
    \\
     \includegraphics[width=0.162\linewidth]{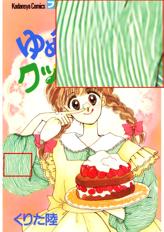}
    \\
    EDSR \cite{20Lim2017CVPRW_EDSR}
    \\
    29.05 / 0.9243
    \end{tabular}
    \end{adjustbox}
    \hspace{-0.55cm}
    \begin{adjustbox}{valign=t}
    \begin{tabular}{c}
    \includegraphics[width=0.162\linewidth]{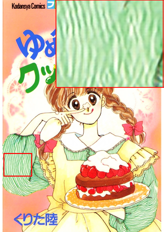} %
    \\
    FSRCNN \cite{16Dong2016ECCV_FSRCNN}
    \\
    26.33 / 0.8440
    \\
     \includegraphics[width=0.162\linewidth]{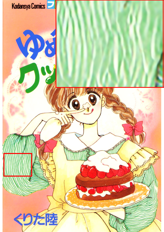} %
    \\
    SRMDNF \cite{50Zhang2018CVPR_SRMDNF}
    \\
    27.51 / 0.8897
    \end{tabular}
    \end{adjustbox}
    \hspace{-0.55cm}
    \begin{adjustbox}{valign=t}
    \begin{tabular}{c}
    \includegraphics[width=0.162\linewidth]{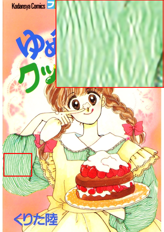} %
    \\
    VDSR \cite{11KimJ2016CVPR_VDSR}
    \\
   27.00 / 0.8744
    \\
     \includegraphics[width=0.162\linewidth]{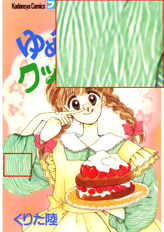}%
    \\
    D-DBPN \cite{27Haris2018CVPR_DBPN}
    \\
    28.27 / 0.9079
    \end{tabular}
    \end{adjustbox}
    \hspace{-0.55cm}
    \begin{adjustbox}{valign=t}
    \begin{tabular}{c}
    \includegraphics[width=0.162\linewidth]{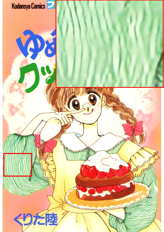}%
    \\
    LapSRN \cite{26LaiWS2017CVPR_LapSRN}
    \\
    26.92 / 0.8752
    \\
     \includegraphics[width=0.162\linewidth]{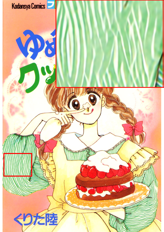}
    \\
    RDN \cite{24Zhang2018CVPR_RDN}
    \\
    28.24 / 0.9121
    \end{tabular}
    \end{adjustbox}
    \hspace{-0.55cm}
    \begin{adjustbox}{valign=t}
    \begin{tabular}{c}
    \includegraphics[width=0.162\linewidth]{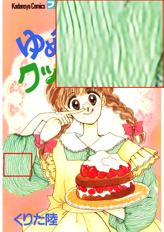}%
    \\
    DRRN \cite{12TaiY2017CVPR_DRRN}
    \\
    27.25 / 0.8826
    \\
     \includegraphics[width=0.162\linewidth]{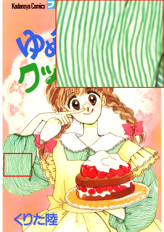}
    \\
    CSFM (Ours)
    \\
   {\bf {29.88} / \bf{0.9379}}
    \end{tabular}
    \end{adjustbox}
   \end{tabular}
\caption{Visual evaluation for a scale factor of $4\times$ on the image ``{\em YumeiroCooking}'' from Manga109.  Our CSFM model can generate finer textures on the sleeves in contrast with other methods which produce the results with severe distortions and heavy blurring artifacts.}
\label{fig: YumeiroCooking_x4}
\end{figure*}


\subsubsection{The Number of FMM Modules  and the Number of CSAR Blocks in each FMM Module}
The capacity of the CSFM network is mainly determined by the number of the FMM modules and the number of the CSAR blocks in each FMM module. In this subsection, we test the effects of  two parameters on image SR. For simplicity, we denote the number of the FMM modules as $M$ and the number of the CSAR blocks as $B$. The network with $m$ modules and $b$ blocks per module is represented as $MmBb$ for short.

Fig.~\ref{fig:The comparisons of different number modules and blocks} shows the results of the PSNR performance (illustrated by different colors according to the color bar on the right) versus two parameters ($M$ and $B$) on the dataset of  BSD100  for a scale factor of $2\times$. We can see that the better performances can be achieved by increasing $M$ or $B$. Since the larger $M$ and $B$ results in a deeper network, the comparisons in Fig.~\ref{fig:The comparisons of different number modules and blocks} suggest that the deeper model is still advantageous. On the other hand, compared with $M4B8$  (achieving 32.212dB on PSNR), $M2B16 $  (obtaining 32.208dB on PSNR) with the same total number of CSAR blocks achieves comparable performance although it has fewer GF nodes for long-term skip connections, and the similar observation can be obtained in the comparison between $M4B16$ and $M8B8$. These results indicate that properly utilizing the limited number of  skip connections does not lose accuracy but reduces the redundancy and computational cost. To effectively exploit long-term skip connections for information persistence as well as control the computational cost, we adopt $M=8$ and $B=16$ as our CSFM model for the next comparison experiments.

\subsection{Comparisons with the State-of-the-arts}
To illustrate the effectiveness of the proposed CSFM network, several state-of-the-art SISR methods, including SRCNN \cite{10Dong2016TPAMI}, FSRCNN \cite{16Dong2016ECCV_FSRCNN}, VDSR \cite{11KimJ2016CVPR_VDSR}, LapSRN \cite{26LaiWS2017CVPR_LapSRN}, DRRN \cite{12TaiY2017CVPR_DRRN}, MemNet \cite{15TaiY2017ICCV_MemNet}, IDN \cite{30Hui2018CVPR_IDN}, EDSR \cite{20Lim2017CVPRW_EDSR}, SRMDNF \cite{50Zhang2018CVPR_SRMDNF}, D-DBPN \cite{27Haris2018CVPR_DBPN} and RDN \cite{24Zhang2018CVPR_RDN}, are compared in terms of quantitative evaluation, visual quality and number of parameters. Since some of existing networks, such as SRCNN, FSRCNN, VDSR, DRRN, MemNet, EDSR and IDN,  did not perform SR on Manga109 dataset, we generate the corresponding results by applying their public trained models to Manga109 dataset for evaluation. In addition, we rebuild the VDSR network in PyTorch with the same network parameters for training and testing as its trained model is not provided.

The quantitative evaluations in the five benchmark datasets for three scale factors ($2\times$, $3\times$, $4\times$) are summarized in TABLE~\ref{tab:Quantitative evaluations of state-of-the-art SR methods}. When compared with MemNet and RDN, both of which introduce persistence memory mechanism via extremely dense skip connections, our CSFM network achieves the highest performance but with fewer skip connections. This indicates that our FMM module with long-term skip connections not only advances the memory block in MemNet \cite{15TaiY2017ICCV_MemNet} and the residual dense block in RDN \cite{24Zhang2018CVPR_RDN} but also reduces the redundancy in the structure of extremely dense connections. Meanwhile, our CSFM model significantly outperforms the remaining methods on all datasets for all upscaling factors, in terms of PSNR and SSIM. Especially, on the challenging dataset Urban100, the proposed CSFM network advances the state-of-the-art (achieved by EDSR or RDN) with the improvement margins of 0.19dB, 0.18dB and 0.14dB on scale factors of $2\times$, $3\times$ and $4\times$ respectively. In addition, more significant improvements earned by the CSFM network are shown on Manga109 dataset, where the proposed CSFM model outperforms EDSR (with highest performance among the prior methods) by the PSNR gains of 0.21dB, 0.32dB and 0.29dB for the $2\times$, $3\times$ and $4\times$ enlargement respectively. These results validate the superiority of the proposed method especially on super-resolving the images with fine structures such as those in Urban100 and Manga109 datasets.

\begin{figure}[tp]
\captionsetup{belowskip=-18pt}
\centering
\includegraphics[width=1.03\linewidth]{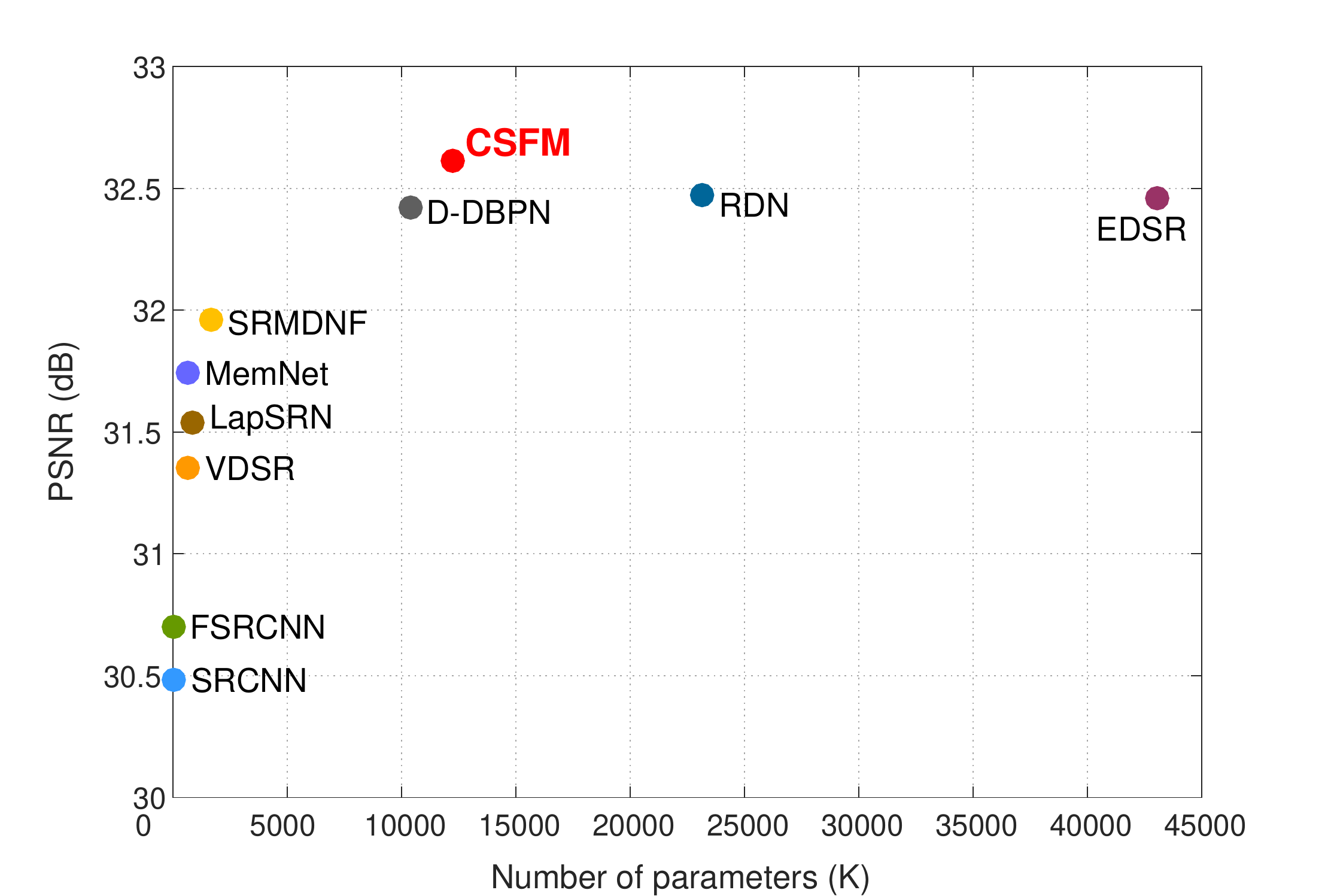}
\caption{ PSNR performance versus number of parameters. The results are evaluated on Set5 dataset for a scale factor of $4\times$. Our CSFM network has a better tradeoff between performance and model size.}
\label{fig: PSNR performance versus number of parameters}
\end{figure}

The visual comparisons of different methods are shown in Fig.~\ref{fig: img092_4x} -- Fig.~\ref{fig: YumeiroCooking_x4}. Thanks to the proposed FMM modules for adaptive multi-level feature-modulation and long-term memory preservation, our proposed CSFM network accurately and clearly reconstructs the stripe patterns, the grid structures, the texture regions and the characters. It is observed that the severe distortions and the noticeable artifacts are contained in the results generated by the prior methods, such as the marked strips on the wall in Fig.~\ref{fig: img092_4x}, the color lines on the balcony in Fig.~\ref{fig: img087_x4} and the  grids on the building in Fig.~\ref{fig: img074_x4}. In contrast, our method avoids the distortions, suppresses the artifacts and generates more faithful results.  Besides, in Fig.~\ref{fig: img076_x4},  Fig.~\ref{fig: YumeiroCooking_x4} and Fig.~\ref{fig: PsychoStaff_x4}, only our method is able to recover more accurate textures and more recognizable characters, while other methods suffer from much information loss and heavy blurring artifacts. The above visual comparisons demonstrate the powerful representational ability of our CSFM network as well.

We also compare the tradeoff between the performance and the number of network parameters from our CSFM network and existing networks. Fig.~\ref{fig: PSNR performance versus number of parameters} shows the PSNR performances of several models versus the number of parameters, where the results are evaluated with Set5 dataset for $4\times$  upscaling factor. We can see that our CSFM network significantly outperforms the relatively small models. Furthermore, compared with EDSR and RDN, our CSFM network achieves higher PSNR but with 72\% and 47\% fewer parameters respectively. These comparisons indicate that our model has a better tradeoff between performance and model size.

\section{Conclusion}
In this paper, we propose a channel-wise and spatial feature modulation (CSFM) network for modeling the process of single image super-resolution, where stacked feature-modulation memory (FMM) modules with the densely connected structure effectively improve its discriminative learning ability and make it concentrate on the worthwhile information. The FMM module consists of a chain of cascaded channel-wise and spatial attention residual (CSAR) blocks and a gated fusion (GF) node. The CSAR block is constructed by incorporating the channel-wise attention and spatial attention into the residual block and utilized to modulate the residual features in a global-and-local way. Further, when a sequence of CSAR blocks are cascaded in the FMM module, two types of attention can be jointly applied to multi-level features and then more informative features can be captured. Meanwhile, The GF node, designed via introducing the gating mechanism and for establishing  long-term skip connections among the FMM modules, can help to maintain long-term information and enhance information flow. Comprehensive evaluations on benchmark datasets demonstrate better performance of our CSFM network in terms of quantitative and qualitative measurements.

{
\bibliographystyle{IEEEtran}
\bibliography{CSFM}

\begin{thebibliography}{10}
\providecommand{\url}[1]{#1}
\csname url@samestyle\endcsname
\providecommand{\newblock}{\relax}
\providecommand{\bibinfo}[2]{#2}
\providecommand{\BIBentrySTDinterwordspacing}{\spaceskip=0pt\relax}
\providecommand{\BIBentryALTinterwordstretchfactor}{4}
\providecommand{\BIBentryALTinterwordspacing}{\spaceskip=\fontdimen2\font plus
\BIBentryALTinterwordstretchfactor\fontdimen3\font minus
  \fontdimen4\font\relax}
\providecommand{\BIBforeignlanguage}[2]{{%
\expandafter\ifx\csname l@#1\endcsname\relax
\typeout{** WARNING: IEEEtran.bst: No hyphenation pattern has been}%
\typeout{** loaded for the language `#1'. Using the pattern for}%
\typeout{** the default language instead.}%
\else
\language=\csname l@#1\endcsname
\fi
#2}}
\providecommand{\BIBdecl}{\relax}
\BIBdecl

\bibitem{1Freeman2000IJCV}
W.~T. Freeman, E.~C. Pasztor, and O.~T. Carmichael, ``Learning low-level
  vision,'' \emph{Int. J. Comput. Vis.}, vol.~40, no.~1, pp. 25--47, Oct. 2000.

\bibitem{2Polatkan2015TPAMI}
G.~Polatkan, M.~Zhou, L.~Carin, and D.~Blei, ``A {Bayesian} non-parametric
  approach to image super-resolution,'' \emph{IEEE Trans. Pattern Anal. Mach.
  Intell.}, vol.~37, no.~2, pp. 346--358, Feb. 2015.

\bibitem{3Chang2004CVPR}
H.~Chang, D.-Y. Yeung, and Y.~Xiong, ``Super-resolution through neighbor
  embedding,'' in \emph{Proc. IEEE Conf. Comput. Vis. Pattern Recognit.
  (CVPR)}, Jun./Jul. 2004, pp. 275--282.

\bibitem{4Jiang2016TCSVT}
J.~Jiang, R.~Hu, Z.~Wang, Z.~Han, and J.~Ma, ``Facial image hallucination
  through coupled-layer neighbor embedding,'' \emph{IEEE Trans. Circuits Syst.
  Video Technol.}, vol.~26, no.~9, pp. 1674--1684, Sep. 2016.

\bibitem{5Yang2010TIP_Sparse}
J.~Yang, J.~Wright, T.~S. Huang, and Y.~Ma, ``Image super-resolution via sparse
  representation,'' \emph{IEEE Trans. Image Process.}, vol.~19, no.~11, pp.
  2861--2873, Nov. 2010.

\bibitem{6He2013CVPR}
L.~He, H.~Qi, and R.~Zaretzki, ``Beta process joint dictionary learning for
  coupled feature spaces with application to single image super-resolution,''
  in \emph{Proc. IEEE Conf. Comput. Vis. Pattern Recognit. (CVPR)}, Jun. 2013,
  pp. 345--352.

\bibitem{7Timofte2014ACCV_A+}
R.~Timofte, V.~D. Smet, and L.~V. Gool, ``A+: adjusted anchored neighborhood
  regression for fast super-resolution,'' in \emph{Proc. 12th Asian Conf.
  Comput. Vis. (ACCV)}, Nov. 2014, pp. 111--126.

\bibitem{8Hu2016TIP_SERF}
Y.~Hu, N.~Wang, D.~Tao, X.~Gao, and X.~Li, ``{SERF}: a simple, effective,
  robust, and fast image super-resolver from cascaded linear regression,''
  \emph{IEEE Trans. Image Process.}, vol.~25, no.~9, pp. 4091--4102, Sep. 2016.

\bibitem{Huang2017TCSVT}
J.~Huang and W.~Siu, ``Learning hierarchical decision trees for single-image
  super-resolution,'' \emph{IEEE Trans. Circuits Syst. Video Technol.},
  vol.~27, no.~5, pp. 937--950, May. 2017.

\bibitem{9Schulter2015CVPR_SRforests}
S.~Schulter, C.~Leistner, and H.~Bischof, ``Fast and accurate image upscaling
  with super-resolution forests,'' in \emph{Proc. IEEE Conf. Comput. Vis.
  Pattern Recognit. (CVPR)}, Jun. 2015, pp. 3791--3799.

\bibitem{44Huang2015CVPRSelfExp}
J.-B. Huang, A.~Singh, and N.~Ahuja, ``Single image super-resolution from
  transformed self-exemplars,'' in \emph{IEEE Conf. Comput. Vis. Pattern
  Recognit. (CVPR)}, Jun. 2015, pp. 5197--5206.

\bibitem{20Lim2017CVPRW_EDSR}
B.~Lim, S.~Son, H.~Kim, S.~Nah, and K.~M. Lee, ``Enhanced deep residual
  networks for single image super-resolution,'' in \emph{Proc. IEEE Conf.
  Comput. Vis. Pattern Recognit. Workshops (CVPRW)}, Jul. 2017, pp. 136--144.

\bibitem{24Zhang2018CVPR_RDN}
Y.~Zhang, Y.~Tian, Y.~Kong, B.~Zhong, and Y.~Fu, ``Residual dense network for
  image super-resolution,'' in \emph{Proc. IEEE Conf. Comput. Vis. Pattern
  Recognit. (CVPR)}, Jun. 2018, pp. 2472--2481.

\bibitem{10Dong2016TPAMI}
C.~Dong, C.~C. Loy, K.~He, and X.~Tang, ``Image super-resolution using deep
  convolutional networks,'' \emph{IEEE Trans. Pattern Anal. Mach. Intell.},
  vol.~38, no.~2, pp. 295--307, Feb. 2016.

\bibitem{11KimJ2016CVPR_VDSR}
J.~Kim, J.~K. Lee, and K.~M. Lee, ``Accurate image super-resolution using very
  deep convolutional networks,'' in \emph{Proc. IEEE Conf. Comput. Vis. Pattern
  Recognit. (CVPR)}, Jun. 2016, pp. 1646--1654.

\bibitem{12TaiY2017CVPR_DRRN}
Y.~Tai, J.~Yang, and X.~Liu, ``Image super-resolution via deep recursive
  residual network,'' in \emph{Proc. IEEE Conf. Comput. Vis. Pattern Recognit.
  (CVPR)}, Jul. 2017, pp. 3147--3155.

\bibitem{13KimJ2016CVPR_DRCN}
{J. Kim, and J. K. Lee, and K. M. Lee}, ``Deeply-recursive convolutional
  network for image super-resolution,'' in \emph{Proc. IEEE Conf. Comput. Vis.
  Pattern Recognit. (CVPR)}, Jun. 2016, pp. 1637--1645.

\bibitem{14Mao2016NIPS_RED}
X.-J. Mao, C.~Shen, and Y.-B. Yang, ``Image restoration using very deep
  convolutional encoder-decoder networks with symmetric skip connections,'' in
  \emph{Proc. Adv. Neural Inf. Process. Syst. (NIPS)}, Dec. 2016, pp.
  2802--2810.

\bibitem{15TaiY2017ICCV_MemNet}
Y.~Tai, J.~Yang, X.~Liu, and C.~Xu, ``{MemNet:} a persistent memory network for
  image restoration,'' in \emph{Proc. IEEE Int. Conf. Comput. Vis. (ICCV)},
  Oct. 2017, pp. 4549--4557.

\bibitem{16Dong2016ECCV_FSRCNN}
C.~Dong, C.~C. Loy, and X.~Tang, ``Accelerating the super-resolution
  convolutional neural network,'' in \emph{Proc. Eur. Conf. Comput. Vis.
  (ECCV)}, Oct. 2016, pp. 391--407.

\bibitem{17Shi2016CVPR_Sub-Pixel}
W.~Shi, J.~Caballero, F.~Huszar, J.~Totz, A.~P. Aitken, R.~Bishop, D.~Rueckert,
  and Z.~Wang, ``Real-time single image and video super-resolution using an
  efficient sub-pixel convolutional neural network,'' in \emph{Proc. IEEE Conf.
  Comput. Vis. Pattern Recognit. (CVPR)}, Jun. 2016, pp. 1874--1883.

\bibitem{18LedigC2017CVPR_SRResNet}
C.~Ledig, L.~Theis, F.~Huszar, J.~Caballero, A.~P. Aitken, A.~Tejani, J.~Totz,
  Z.~Wang, and W.~Shi, ``Photo-realistic single image super-resolution using a
  generative adversarial network,'' in \emph{Proc. IEEE Conf. Comput. Vis.
  Pattern Recognit. (CVPR)}, Jul. 2017, pp. 105--114.

\bibitem{19He2016CVPR_ResnetIR}
K.~He, X.~Zhang, S.~Ren, and J.~Sun, ``Deep residual learning for image
  recognition,'' in \emph{Proc. IEEE Conf. Comput. Vis. Pattern Recognit.
  (CVPR)}, Jun. 2016, pp. 770--778.

\bibitem{22Huang2017CVPR_DenseNet}
G.~Huang, Z.~Liu, K.~Q. Weinberger, and L.~van~der Maaten, ``Densely connected
  convolutional networks,'' in \emph{Proc. IEEE Conf. Comput. Vis. Pattern
  Recognit. (CVPR)}, Jul. 2017, pp. 4700--4708.

\bibitem{23Tong2017ICCV_DenseSR}
T.~Tong, G.~Li, X.~Liu, and Q.~Gao, ``Image super-resolution using dense skip
  connections,'' in \emph{Proc. IEEE Int. Conf. Comput. Vis. (ICCV)}, Oct.
  2017, pp. 4799--4807.

\bibitem{25Wang2018CVPRW_ProSR}
Y.~Wang, F.~Perazzi, B.~McWilliams, A.~Sorkine-Hornung, O.~Sorkine-Hornung, and
  C.~Schroers, ``A fully progressive approach to single-image
  super-resolution,'' in \emph{Proc. IEEE Conf. Comput. Vis. Pattern Recognit.
  Workshops (CVPRW)}, Jun. 2018, pp. 977--986.

\bibitem{26LaiWS2017CVPR_LapSRN}
W.-S. Lai, J.-B. Huang, N.~Ahuja, and M.-H. Yang, ``Deep {Laplacian} pyramid
  networks for fast and accurate super-resolution,'' in \emph{Proc. IEEE Conf.
  Comput. Vis. Pattern Recognit. (CVPR)}, Jul. 2017, pp. 624--632.

\bibitem{He2018TCSVT}
Z.~He, S.~Tang, J.~Yang., Y.~Cao, M.~Y. Yang, and Y.~Cao, ``Cascaded deep
  networks with multiple receptive fields for infrared image
  super-resolution,'' \emph{IEEE Trans. Circuits Syst. Video Technol.}, 2018.

\bibitem{27Haris2018CVPR_DBPN}
M.~Haris, G.~Shakhnarovich, and N.~Ukita, ``Deep back-projection networks for
  super-resolution,'' in \emph{Proc. IEEE Conf. Comput. Vis. Pattern Recognit.
  (CVPR)}, Jun. 2018, pp. 1664--1673.

\bibitem{28Han2018arXiv_DSRN}
S.~Zagoruyko and N.~Komodakis, ``Image super-resolution via dual-state
  recurrent networks,'' \emph{arXiv: 1805.02704}, May. 2018.

\bibitem{30Hui2018CVPR_IDN}
Z.~Hui, X.~Wang, and X.~Gao, ``Fast and accurate single image super-resolution
  via information distillation network,'' in \emph{Proc. IEEE Conf. Comput.
  Vis. Pattern Recognit. (CVPR)}, Jun. 2018, pp. 723--731.

\bibitem{29Kim2018CVPRW_EUSR}
J.-H. Kim and J.-S. Lee, ``Deep residual network with enhanced upscaling module
  for super-resolution,'' in \emph{Proc. IEEE Conf. Comput. Vis. Pattern
  Recognit. Workshops (CVPRW)}, Jun. 2018, pp. 913--921.

\bibitem{31Mansimov2016ICLR_CaptionAtten}
E.~Mansimov, E.~Parisotto, J.~L. Ba, and R.~Salakhutdinov, ``Generating images
  from captions with attention,'' in \emph{Int. Conf. Learn. Rep. (ICLR)}, May.
  2016, pp. 1--4.

\bibitem{32Xu2015ICML_CaptionGenera}
K.~Xu, J.~Ba, R.~Kiros, K.~Cho, A.~Courville, R.~Salakhutdinov, R.~S. Zemel,
  and Y.~Bengio, ``Show, attend and tell: Neural image caption generation with
  visual attention,'' in \emph{Int. Conf. Mach. Learn. (ICML)}, Jul. 2015, pp.
  2048--2057.

\bibitem{33Chen2017CVPR_SCA-CNN}
L.~Chen, H.~Zhang, J.~Xiao, L.~Nie, J.~Shao, W.~Liu, and T.-S. Chua, ``Sca-cnn:
  Spatial and channel-wise attention in convolutional networks for image
  captioning,'' in \emph{Proc. IEEE Conf. Comput. Vis. Pattern Recognit.
  (CVPR)}, Jul. 2017, pp. 5659--5667.

\bibitem{34Hu2018CVPR_SEnet}
J.~Hu, L.~Shen, and G.~Sun, ``Squeeze-and-excitation networks,'' in \emph{Proc.
  IEEE Conf. Comput. Vis. Pattern Recognit. (CVPR)}, Jun. 2018, pp. 7132--7141.

\bibitem{35Wang2017CVPR_ResidualAttenClass}
F.~Wang, M.~Jiang, C.~Qian, S.~Yang, C.~Li, H.~Zhang, X.~Wang, and X.~Tang,
  ``Residual attention network for image classification,'' in \emph{Proc. IEEE
  Conf. Comput. Vis. Pattern Recognit. (CVPR)}, Jul. 2017, pp. 3156--3164.

\bibitem{21Zhang2018ECCV_RCAN}
Y.~Zhang, K.~Li, K.~Li, L.~Wang, B.~Zhong, and Y.~Fu, ``Image super-resolution
  using very deep residual channel attention networks,'' in \emph{Proc. Eur.
  Conf. Comput. Vis. (ECCV)}, Sept. 2018, pp. 1--16.

\bibitem{36Wang2018CVPR_SFTSR}
X.~Wang, K.~Yu, C.~Dong, and C.~C. Loy, ``Recovering realistic texture in image
  super-resolution by deep spatial feature transform,'' in \emph{Proc. IEEE
  Conf. Comput. Vis. Pattern Recognit. (CVPR)}, Jun. 2018, pp. 606--615.

\bibitem{37Xu2016ECCV_QuestAnswer}
H.~Xu and K.~Saenko, ``Ask, attend and answer: Exploring question guided
  spatial attention for visual question answering,'' in \emph{Proc. Eur. Conf.
  Comput. Vis. (ECCV)}, Oct. 2016, pp. 451--466.

\bibitem{38Chen2017NIPS_DualPN}
Y.~Chen, J.~Li, H.~Xiao, X.~Jin, S.~Yan, and J.~Feng, ``Dual path networks,''
  in \emph{Proc. Adv. Neural Inf. Process. Syst. (NIPS)}, Dec. 2017, pp. 1--9.

\bibitem{39Nair2010ICML_ReLU}
V.~Nair and G.~Hinton, ``Rectified linear units improve restricted boltzmann
  machines,'' in \emph{Int. Conf. Mach. Learn. (ICML)}, Jun. 2010, pp.
  807--814.

\bibitem{40Itti2001Neurosci}
L.~Itti and C.~Koch, ``Computational modelling of visual attention,''
  \emph{Nat. Rev. Neurosci.}, vol.~2, no.~3, pp. 194--203, Mar. 2001.

\bibitem{41Bevilacqua2012BMVC}
M.~Bevilacqua, A.~Roumy, C.~Guillemot, and M.-L.~A. Morel, ``Low-complexity
  single-image super-resolution based on nonnegative neighbor embedding,'' in
  \emph{Proc. 23rd British Mach. Vis. Conf. (BMVC)}, Sep. 2012, pp.
  135.1--135.10.

\bibitem{42Zeyde2010ICCS}
R.~Zeyde, M.~Elad, and M.~Protter, ``On single image scale-up using
  sparse-representations,'' in \emph{Proc. 7th Int. Conf. Curves Surfaces},
  Jun. 2010, pp. 711--730.

\bibitem{43ArbelaezP2011TPAMI}
P.~Arbel{\'{a}}ez, M.~Maire, C.~Fowlkes, and J.~Malik, ``Contour detection and
  hierarchical image segmentation,'' \emph{IEEE Trans. Pattern Anal. Mach.
  Intell.}, vol.~33, no.~5, pp. 898--916, May. 2011.

\bibitem{45Matsui2017MultiTollAppli}
Y.~Matsui, K.~Ito, Y.~Aramaki, A.~Fujimoto, T.~Ogawa, T.~Yamasaki, and
  K.~Aizawa, ``Sketch-based manga retrieval using manga109 dataset,''
  \emph{Multimedia Tools Appl.}, vol.~76, no.~20, pp. 21\,811--21\,838, Oct.
  2017.

\bibitem{46Timofte2017CVPRW_NTIRE}
R.~Timofte, E.~Agustsson, L.~V. Gool, M.-H. Yang, L.~Zhang, B.~Lim, and et~al.,
  ``{NTIRE} 2017 challenge on single image super-resolution: methods and
  results,'' in \emph{Proc. IEEE Conf. Comput. Vis. Pattern Recognit. Workshops
  (CVPRW)}, Jul. 2017, pp. 1110--1121.

\bibitem{47Wang2004TIP_SSIM}
Z.~Wang, A.~C. Bovik, H.~R. Sheikh, and E.~P. Simoncelli, ``Image quality
  assessment: from error visibility to structural similarity,'' \emph{IEEE
  Trans. Image Process.}, vol.~13, no.~4, pp. 600--612, Apr. 2004.

\bibitem{48Kingma2015ICLR_Adam}
D.~P. Kingma and J.~Ba, ``Adam: A method for stochastic optimization,'' in
  \emph{Int. Conf. Learn. Rep. (ICLR)}, May. 2015, pp. 1--13.

\bibitem{49Paszke2017NIPSWAutodiff_Pytorch}
A.~Paszke, S.~Gross, S.~Chintala, G.~Chanan, E.~Yang, Z.~DeVito, Z.~Lin,
  A.~Desmaison, L.~Antiga, and A.~Lerer, ``Automatic differentiation in
  {PyTorch},'' in \emph{Proc. Adv. Neural Inf. Process. Syst. Workshop
  Autodiff}, Dec. 2015, pp. 1--4.

\bibitem{50Zhang2018CVPR_SRMDNF}
K.~Zhang, W.~Zuo, and L.~Zhang, ``Learning a single convolutional
  super-resolution network for multiple degradations,'' in \emph{Proc. IEEE
  Conf. Comput. Vis. Pattern Recognit. (CVPR)}, Jun. 2018, pp. 3262--3271.

\end{thebibliography}
}

\end{document}